\documentclass{clv3}
\pdfoutput=1

\usepackage{hyperref}
\usepackage{url}

\usepackage{xcolor}
\definecolor{darkblue}{rgb}{0, 0, 0.5}
\hypersetup{colorlinks=true,citecolor=darkblue, linkcolor=darkblue, urlcolor=darkblue}

\usepackage{graphicx}
\usepackage{tikz}
\usetikzlibrary{trees,positioning,shapes,shadows,arrows}
\usepackage{caption}
\usepackage{subcaption}

\usepackage{CJKutf8}

\usepackage{amsmath}
\usepackage{lettrine}

\usepackage[textsize=footnotesize]{todonotes}
\usepackage{comment}
\usepackage{enumitem}
\usepackage{cleveref}
\crefformat{section}{\S#2#1#3}

\usepackage{booktabs}

\issue{1}{1}{2016}

\runningtitle{Survey of Methods for Low-resource Machine Translation}

\runningauthor{Barry Haddow}

\begin{document}

\title{Survey of Low-Resource Machine Translation}

\author{Barry Haddow}
\affil{University of Edinburgh}

\author{Rachel Bawden}
\affil{Inria, France}

\author{Antonio Valerio Miceli Barone}
\affil{University of Edinburgh}

\author{Jind\v{r}ich Helcl}
\affil{University of Edinburgh}

\author{Alexandra Birch\thanks{10 Crichton Street, EH89AB, UK, a.birch@ed.ac.uk}}
\affil{University of Edinburgh}

\maketitle

\begin{abstract}
We present a survey covering the state of the art in low-resource machine translation research. 
There are currently around 7000 languages spoken in the world and almost all language pairs lack significant resources for training machine translation models. 
There has been increasing interest in research addressing the challenge of producing useful translation models when very little translated training data is available. 
We present a  summary of this topical research field and provide a description of the techniques evaluated by researchers in several recent shared tasks in low-resource MT. 
\end{abstract}


\section{Introduction}

Current machine translation (MT) systems have reached the stage where researchers are now debating whether or not they can rival human translators in performance \cite{DBLP:journals/corr/abs-1803-05567,2018arXiv180807048L,toral-etal-2018-attaining,popel2020nature}. 
However these MT systems are typically trained on data sets consisting of tens or even hundreds of millions of parallel sentences. 
Data sets of this magnitude are only available for a small number of highly resourced language pairs (typically English paired with some European languages, Arabic and Chinese). 
The reality is that for the vast majority of language pairs in the world the amount of data available is extremely limited, or simply non-existent. 

 \begin{table}[h]
 \small
 \begin{center}
\begin{tabular}{lrr}
\toprule
 Language Pair & Speakers (approx) & Parallel Sentences   \\
\midrule
 English--French & 267M & 280M \\
 English--Myanmar & 30M & 0.7M \\
 English--Fon & 2M & 0.035M \\
\bottomrule
\end{tabular}
\caption{Examples of language pairs with different levels of resources. The number of speakers is obtained from Ethnologue, and the parallel sentence counts are from Opus.}
\label{tab:example}
\end{center}
\end{table}

In Table~\ref{tab:example} we select three language pairs that display a range of resource levels.  We show the estimated number of first language speakers\footnote{from Ethnologue: \url{https://www.ethnologue.com/}} of the non-English language, together with the number of parallel sentences 
available in Opus \citep{TIEDEMANN12.463}, the largest collection of publicly available translated 
data.\footnote{The Opus counts were accessed in September 2021.}$^,$\footnote{There are of course parallel corpora that are not in Opus, and so these numbers remain an approximation, but we believe that due to its size and coverage, Opus counts can be taken as indicative of the total available parallel data.}
Although there is typically a correlation between speaker numbers and size of available resources, there are many exceptions where either widely spoken languages have little parallel data, or languages with very small speaker numbers are richly resourced (mainly European languages). 
For one of the world's most spoken languages, French, there are nearly 280 million parallel sentences of English--French in OPUS. 
However when we search for English--Myanmar, we find only around 700,000 parallel sentences, despite Myanmar having tens of millions of speakers. If we consider Fon, which has around 2 million speakers, then we find far fewer parallel sentences, only 35,000\footnote{To find further resources for Fon,
we could consider Fon--French \cite{emezue-dossou-2020-ffr}, where there is more parallel than for 
Fon--English, although the cleaned corpus still has fewer than 55k sentence pairs}.  
Developing MT systems for these three language pairs will 
require very different techniques.

The lack of parallel data for most language pairs is only one part of the problem. Existing data is often noisy or from a very specialised domain. Looking at the resources that are available for Fon--English, we see that the only corpus available in Opus is extracted from Jehovah's Witness publications \citep[JW300]{agic-vulic-2019-jw300}\footnote{Unfortunately the JW300 corpus is no longer available to researchers, since the publishers have asked for it to be withdrawn for breach of copyright.}. For many language pairs, the only corpora available are those derived from religious sources (e.g.~Bible, Koran) or from IT localisation data (e.g.~from open-source projects such as GNOME and Kubuntu). Not only is such data likely to be in a very different domain from the text that we would like to translate, but such large-scale multilingual automatically extracted corpora are often of poor quality \citep{caswell2021-quality-at-a-glance} 
and this problem is worse for low-resource language pairs. 
This means that low-resource language pairs suffer from multiple compounding problems: lack of data, out-of-domain data and noisy data. And the difficulty is not just with parallel data, low-resource languages often lack good linguistic tools, and even  basic tools like language identification do not exist or are not reliable.

Partly in response to all these challenges, there has been an increasing interest in the research community in exploring more diverse languages, and language pairs that do not include English. 
This survey paper presents a high-level summary of approaches to low-resource MT, with focus on neural machine translation (NMT) techniques, which should be useful for researchers interested in this broad and rapidly evolving field.
There are currently a number of other survey papers in related areas, for example a survey of monolingual data in low-resource NMT~\cite{gibadullin2019survey} and a survey of multilingual NMT~\cite{Dabre2020-bb}. There have also been two very recent surveys of low-resource MT, which have been written concurrently with this survey~\cite{Ranathunga2021-qm,wang2021survey}. Our survey aims to provide the broadest coverage of existing research in the field and we also contribute an extensive overview of the tools and techniques validated across 18 low-resource shared tasks that ran between 2018 and 2021. 

One of the challenges of surveying the literature on low-resource MT is how to define what a low-resource language pair is. This is hard, because ``resourced-ness'' is a continuum and any criterion must be arbitrary. We also note that the definition of low-resource can change over time. We could crawl more parallel data, or we could find better ways of using related language data or monolingual data which means that some language pairs are no longer so resource-poor. 
 We maintain that for research to be considered to be on ``low-resource MT'', there should be some way in which 
the research should either aim to understand the implications of the lack of data, or propose methods for overcoming the lack of data. 
We do not take a strict view of what to include in this survey though; if the authors consider that they are studying low-resource MT, then that is sufficient. 
We do feel however that it is important to distinguish between simulated low-resource settings (where a limited amount of data from otherwise high-resource language pairs is used) and genuinely low-resource languages (where additional difficulties apply).
We also discuss some papers that do not explicitly consider low-resource MT but which present important techniques and we mention methods that we think have the potential to improve low-resource MT. 
Even though there has been a lot of interest recently in low-resource NLP, the field is limited to languages where some textual data is freely available. This means that so far low-resource MT has only considered 
100-odd languages, and there is a long tail of languages that is still unexplored.

\begin{figure}[ht!]
     \centering
     \resizebox{\textwidth}{!}{\tikzset{
  basic/.style  = {draw,text width=3.5cm, font=\sffamily, rectangle},
  root/.style   = {basic, draw=none, rounded corners=2pt, thin, align=center, fill=white},
  level-2/.style = {basic, draw, thin, rounded corners=6pt, align=center, text width=2.5cm, fill=teal!40, scale=1},
  level-22/.style= {level-2, fill=teal!20, text width=2.5cm},
  level-3/.style = {basic, thin, align=left, fill=white, text width=2.7cm},
  level-33/.style = {level-3, scale=0.9, draw=none, align=left, yshift=8pt, text width=2cm}
}

\begin{tikzpicture}[
  level 1/.style={sibling distance=20em, level distance=5em},
  edge from parent/.style={->,solid,black,thick,sloped,draw}, 
  edge from parent path={(\tikzparentnode.south) -- (\tikzchildnode.north)},
  >=latex, node distance=1.2cm, edge from parent fork down,
  level 2/.style={sibling distance=10em, level distance=5em}]

\node[root] {\large\textbf{Low-resource MT}}
  child {node[level-2,yshift=-50pt] (c1) {§\ref{sec:data}~\textbf{Data collection}}}
  child {node[level-2] (c2) {\textbf{Data exploitation}}
    child {node[level-22] (c21) {§\ref{sec:monolingual}~Monolingual data}}
    child {node[level-22] (c22) {§\ref{sec:multilingual}~Multilingual data}}
    child {node[level-22] (c23) {§\ref{sec:linguistic}~Other resources}}
  }
  child {node[level-2, yshift=-50pt] (c3) {§\ref{sec:ml}~\textbf{Model \\choices}}}
  ;
  

\begin{scope}[every node/.style={level-3}]
\node [below of = c1, xshift=10pt] (c11) {§\ref{sec:data-existing}~Existing data};
\node [below of = c11] (c12) {§\ref{sec:data-crawl}~Web crawling};

\node [below of = c12] (c13) {§\ref{sec:data-other}~Data creation};
\node [below of = c13] (c14) {§\ref{sec:data-test}~Test data};

\node [below of = c21, xshift=10pt] (c211) {§\ref{sec:mono-external-lm}~External LMs};
\node [below of = c211] (c212) {§\ref{sec:mono-via-synth}~Parallel data synthesis};
\node [below of = c212, xshift=10pt, level-33] (c2121) {\small\textit{Backtranslation and variants}};
\node [below of = c2121, level-33] (c2122) {\small\textit{Unsupervised MT}};
\node [below of = c2122, level-33] (c2123) {\small\textit{Modifying parallel data}};
\end{scope}
\begin{scope}[every node/.style={level-3}]
\node [below of = c2123,xshift=-10pt] (c213) {§\ref{sec:mono-via-xfer}~Transfer learning};
\node [below of = c213,xshift=10pt, level-33] (c2131) {\small\textit{Pretrained embeddings}};
\node [below of = c2131, level-33] (c2132) {\small\textit{Pretraining with LMs}};
\end{scope}
\begin{scope}[every node/.style={level-3}]

\node [below of = c22, xshift=10pt] (c221) {§\ref{sec:multi-transfer}~Transfer learning};
\node [below of = c221] (c222) {§\ref{sec:multi-model}~Multilingual models};
\node [below of = c222] (c223) {§\ref{sec:multi-pretraining}~Large-scale pre-training};

\node [below of = c23, xshift=10pt] (c231) {§\ref{sec:ling:tools-resources}~Linguistic tools};
\node [below of = c231, xshift=10pt, level-33] (c2311) {\small\textit{Morphological segmentation}};
\node [below of = c2311, level-33] (c2312) {\small\textit{Factored models}};
\node [below of = c2312, level-33] (c2313) {\small\textit{Multi-task learning}};

\node [below of = c2313, xshift=-10pt] (c232) {§\ref{sec:ling:lexicons}~Bilingual lexicons};

\node [below of = c3, xshift=10pt] (c31) {§\ref{sec:ml:meta}~Meta-learning};
\node [below of = c31] (c32) {§\ref{sec:ml:latent}~Latent variables};
\node [below of = c32] (c33) {§\ref{sec:ml:train}~Alternative training};
\node [below of = c33] (c34) {§\ref{sec:ml:inference}~Alternative inference};
\node [below of = c34] (c35) {§\ref{sec:ml:rulebased}~Rule-based};

\end{scope}


\foreach \value in {1,2,3,4}
  \draw[->] (c1.192) |- (c1\value.west);

\foreach \value in {1,...,3}
  \draw[->] (c21.192) |- (c21\value.west);
  
\foreach \value in {1,...,3}
  \draw[->] (c212.205) |- (c212\value.west);
  
  \foreach \value in {1,...,2}
  \draw[->] (c213.205) |- (c213\value.west);

\foreach \value in {1,...,3}
  \draw[->] (c22.192) |- (c22\value.west);

\foreach \value in {1,...,2}
  \draw[->] (c23.192) |- (c23\value.west);
  
\foreach \value in {1,...,3}
  \draw[->] (c231.205) |- (c231\value.west);

\foreach \value in {1,...,5}
  \draw[->] (c3.192) |- (c3\value.west);

\end{tikzpicture}
}
     \caption{Overview of research methods covered in this survey. }
     \label{fig:lr-hierarchy}
 \end{figure}
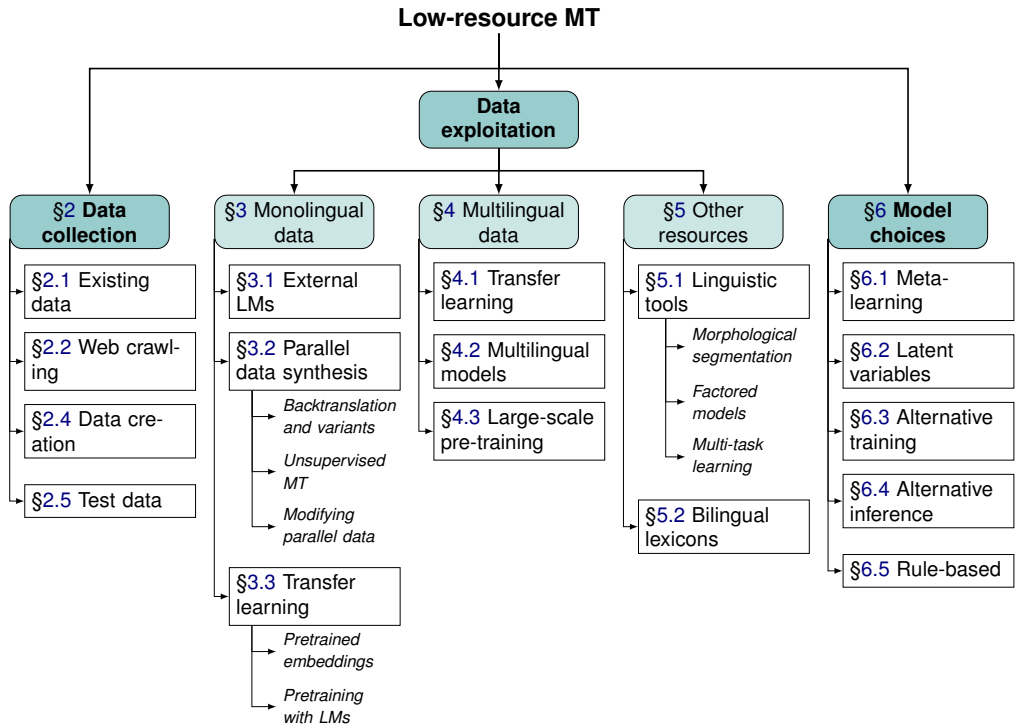

In Figure~\ref{fig:lr-hierarchy} we show how we structure the diverse research methods addressing low-resource MT, and this paper follows this structure. 
We start the survey by looking at work that aims to increase the amount and quality of parallel and monolingual data available for low-resource MT (Section~\ref{sec:data}).
We then look at work that uses other types of data: monolingual data (Section~\ref{sec:monolingual}), parallel data from other language pairs (Section~\ref{sec:multilingual}), and other
types of linguistic data (Section~\ref{sec:linguistic}).  
Another avenue of important research is how to make better use of existing, limited resources through better training or modelling (Section~\ref{sec:ml}). 
In Section~\ref{sec:eval} we pause our discussion on methods to improve low-resource MT, to consider how to evaluate
these improvements.  In our final section, we look at efforts in the community to build research capacity through shared tasks and language-specific collectives (Section~\ref{sec:cases}), providing a practical summary of commonly used approaches and other techniques often used by top-performing shared task systems. 

This survey aims to provide researchers and practitioners interested in low-resource MT with an overview of the area, and we hope it will be especially useful with those that are new to low-resource MT, and looking to quickly assimilate the recent research directions. We assume that our readers have prior knowledge of MT techniques and are already familiar with basic concepts, including the main architectures used. We therefore do not redefine them in this survey and refer the interested reader to other resources such as \citep{koehn_2020}.



\section{Data Sources}
\label{sec:data}

The first consideration when applying data-driven MT to a new language pair is to determine what
data resources are already available. In this section we discuss commonly used data sets and how to extract more data, especially parallel data, for low-resource languages. 
  A recent case-study ~\cite{hasan-etal-2020-low} has shown  how carefully targeted data gathering can lead to clear MT improvements in a low-resource language pair (in this case, Bengali--English).
Data is arguably the most important factor in our success at modelling translation and we encourage readers to consider data set creation and curation as important areas for future work.

\subsection{Searching Existing Data Sources}\label{sec:data-existing}

The largest range of freely available parallel data is found on Opus\footnote{\url{http://opus.nlpl.eu/}} \citep{TIEDEMANN12.463}, which hosts parallel corpora covering more than 500 different languages and variants. Opus collects contributions of parallel corpora from many sources in one convenient website, and provides tools for downloading corpora and metadata.


Monolingual corpora are also useful, although not nearly as valuable as parallel data. There have been a few efforts to extract text from the CommonCrawl\footnote{\url{https://commoncrawl.org/}} collection of websites, and these generally have broad language coverage. The first extraction effort was by \citet{Buck-commoncrawl}, 
although more recent efforts such as Oscar\footnote{\url{https://oscar-corpus.com/}} \citep{ortizsuarez:hal-02148693},  CC100 \cite{conneau-etal-2020-unsupervised} and mc4\footnote{\url{https://huggingface.co/datasets/mc4}} \citep{2019t5} have focused on cleaning the data, and making it easier to access.
A  much smaller (but generally less noisy) corpus of monolingual news\footnote{\url{http://data.statmt.org/news-crawl/}} is updated annually for the WMT shared tasks \citep{barrault-EtAl:2020:WMT1}, and currently covers 59 languages. For languages not covered in any of these data sets, wikipedia currently has text in over 300 languages, although many language sections are quite small. 

\subsection{Web-crawling for Parallel Data}\label{sec:data-crawl}

Once freely available sources of parallel data have been exhausted, one avenue for improving low-resource NMT is to obtain more parallel data by web-crawling. 
There is a large amount of translated text available on the web, ranging from small-scale tourism websites to large repositories of government documents. Identifying, extracting and sentence-aligning such texts is not straightforward, and researchers have considered many techniques for producing parallel corpora from web data. The links between source texts and their translations are rarely recorded in a consistent way, so techniques ranging from simple heuristics to neural sentence embedding methods are used to extract parallel documents and sentences. 

Paracrawl,\footnote{\url{https://www.paracrawl.eu/}} a recent large-scale open-source crawling effort \citep{banon-etal-2020-paracrawl} has mainly targeted European languages, only a few of which can be considered as low-resource, but it has created some releases for non-European low-resource language pairs, and the crawling 
pipeline is freely available. Related to this are other broad parallel data extraction efforts, where recently developed sentence-embedding based alignment methods \citep{artetxe-schwenk-2019-massively,artetxe-schwenk-2019-margin} were used to extract large parallel corpora in many language pairs from Wikipedia \citep{2019arXiv190705791S} and CommonCrawl
\citep{schwenk2019ccmatrix}. Similar techniques \citep{feng2020languageagnostic} were used to create the largest parallel corpora of Indian languages, Samanantar \citep{samanantar}.

Broad crawling and extraction efforts are good for gathering data from the ``long tail'' of small websites with parallel data, but they tend to be much more effective for high-resource language pairs, because there is more text in these languages, and routine translation of websites is more common for some languages. Focused crawling efforts, where the focus is on particular language pairs, or particular sources, can be more effective for boosting the available data for low-resource language pairs. 
Religious texts are often translated into many different languages, with the texts released under permissive licences to encourage dissemination, so these have been the focus of some recent crawling efforts, such 
as corpora from the Bible \citep{mayer-cysouw-2014-creating,10.1007/s10579-014-9287-y} or the Quran \citep{TIEDEMANN12.463}.
However, a permissive license should not be automatically assumed even for religious publications, for instance a corpus of Jehovah's Witnesses publications \cite{agic-vulic-2019-jw300} was recently withdrawn due to a copyright infringement claim, therefore we recommend to always check the license of any material one intends to crawl and if unclear ask the original publisher for permission. 

In India, government documents are often translated into many of the country's official languages, most of which would be considered low-resource for MT. Recent efforts \cite{haddow2020pmindia, siripragada-etal-2020-multilingual, philip2020revisiting} have been made to gather and align this data, including producing parallel corpora \emph{between} different Indian languages, rather than the typical English-centric parallel corpora. The last of these works uses an iterative procedure, starting with an existing NMT system
for alignment \citep{W11-4624},  and proceeding through rounds of crawling, alignment and retraining, to produce parallel corpora for languages of India.
Further language-specific data releases for low-resource languages, including not only crawled data from MT but annotated also data for other NLP tasks, were recently provided by the Lorelei project \citep{tracey-etal-2019-corpus}, although this data is distributed under restrictive and sometimes costly LDC licences.

\subsection{Low-resource Languages and Web-crawling}\label{sec:data-web}\label{sec:data-low-crawl}

Large-scale crawling faces several problems when targeting low-resource language pairs. Typical crawling pipelines require several stages and make use of resources such as text preprocessors, bilingual dictionaries, sentence-embedding tools and translation systems, all of which may be unavailable or of poor quality for low-resource pairs. Also, parallel sentences are so rare for low-resource language pairs (relative to the size of the web)  that even a small false-positive rate will result in a crawled corpus that is mostly noise (e.g. sentences that are badly aligned, sentences in the wrong language, or fragments of html/javascript).

One of the first stages of any crawling effort is language identification, perhaps thought to be a solved problem with wide-coverage open-source toolkits such as CLD3.\footnote{\url{https://github.com/google/cld3}} However it has been noted \citep{caswellc2020lid} that language identification can perform quite poorly on web-crawled corpora, especially on low-resource languages, where it is affected by class imbalance, similar languages, encoding problems and domain mismatches. Further down the crawling pipeline, common techniques for document alignment and sentence alignment rely on the existence
of translation systems \cite{uszkoreit-EtAl:2010:PAPERS, sennrich-volk-2010-mt,W11-4624} or sentence embeddings \citep{artetxe-schwenk-2019-margin}, which again may not be of sufficient quality in low-resource languages and so we often have to fall back on older, heuristic alignment
techniques \citep{hunalign} (and even this may perform worse if a bilingual lexicon is not
available). The consequence is that the resulting corpora are extremely noisy and require extensive filtering before they can be useful for NMT training.

Filtering of noisy corpora is itself an active area of research, and has been explored in recent shared tasks, which particularly emphasised low-resource settings \citep{koehn-etal-2019-findings,koehn-EtAl:2020:WMT}. In an earlier version
of the task \citep{koehn-etal-2018-findings}, dual conditional cross-entropy \citep[DCCE]{junczys-dowmunt-2018-dual} was found to be very effective for English--German. 
In the 2019 and 2020 editions of the task however, DCCE was much less used possibly indicating that
it is less effective for low-resource and/or distant language pairs. Instead, we see that all participants apply some heuristic filtering (e.g.\ based on language identification and length) and then strong submissions typically used a combination of embedding-based
methods (such as LASER \citep{artetxe-schwenk-2019-massively}, GPT-2 \citep{radford2019language} and YiSi \citep{lo-2019-yisi}) with
feature-based systems such as zipporah \citep{Xu2017ZipporahAF} or bicleaner \citep{prompsit:2018:WMT}. Whilst the feature-based methods are much faster than sentence-embedding based methods, both types of methods require significant effort in transferring to a new language pair, especially if no pre-trained sentence embeddings or other models are available.

The conclusion is that all crawled data sources should be treated with care, especially in low-resource settings as they will inevitably contain errors. A large-scale quality
analysis \cite{caswell2021-quality-at-a-glance} of crawled data has highlighted that many contain incorrect language identification, non-parallel sentences, low
quality texts, as well as offensive language, and these problems can be more acute in low-resource languages.

\subsection{Other Data Sources}\label{sec:data-other}

In addition to mining existing sources of translations, researchers have turned their attention to ways of creating new parallel data. One idea for doing this in a cost-effective manner is crowd-sourcing of translations. \citet{post-etal-2012-constructing} showed this method is effective in collecting a parallel corpus covering several
languages of India. \citet{Pavlick-EtAl-2014:TACL} used crowd-sourcing to collect bilingual dictionaries covering a large selection of languages. Whilst not specifically aimed at MT, the 
Tatoeba\footnote{\url{https://tatoeba.org/}} collection of crowd-sourced translations provides a 
useful resource with broad language coverage. An MT challenge set covering  over 100 language pairs
has been derived from Tatoeba \citep{tiedemann-2020-tatoeba}.


\subsection{Test Sets}\label{sec:data-test}

Obtaining more training data is important, but we should not forget the role of standardised and reliable test sets in improving performance on low-resource translation. Important contributions have come from shared tasks, such as those organised
by the WMT Conference on Machine Translation \citep{bojar-EtAl:2018:WMT1, barrault-etal-2019-findings, barrault-EtAl:2020:WMT1, fraser-2020-findings}, the Workshop on Asian Translation \citep{nakazawa-etal-2018-overview, nakazawa-etal-2019-overview, nakazawa-etal-2020-overview} and the Workshop on Technologies for MT of Low Resource Languages (LoResMT) \citep{ws-2019-technologies,ojha-etal-2020-findings}. 

The test sets in these shared tasks are very useful, but inevitably only cover a small selection of low-resource languages, and usually English is one of the languages in the pair; non-English language pairs are poorly covered. 
A recent initiative towards rectifying this situation is  the FLORES-101 
benchmark \citep{flores101}, which covers  a large number of low-resource languages with multi-parallel test sets, vastly expanding on the original FLORES release \citep{guzman-etal-2019-flores}. Since FLORES-101 consists of the same set of English sentences, translated into 100 other languages, it can also be used for testing translation between non-English pairs. It has the limitation that, for non-English pairs, the two sides are ``translationese'', and not mutual translations of each
other, but there is currently no other data set with the coverage
of FLORES-101.

We summarise the data sets described in this section in Table~\ref{tab:corpora}.

\begin{table}
\centering\small
\resizebox{\textwidth}{!}{
\rotatebox{0}{
\begin{tabular}{p{2.2cm}lp{5.5cm}}
\toprule
Corpus name & URL & Description \\
\midrule
CC100 & {\small \url{http://data.statmt.org/cc-100/}}  & Monolingual data from CommonCrawl (100 languages).  \\
Oscar & {\small \url{https://oscar-corpus.com/}} &  Monolingual data from CommonCrawl (166 languages).\\
mc4 & {\small \url{https://huggingface.co/datasets/mc4}} & Monolingual data from CommonCrawl (108 languages). \\
news-crawl & {\small \url{http://data.statmt.org/news-crawl/}} &  Monolingual news text in 59 languages.\\
\midrule
Opus & {\small \url{https://opus.nlpl.eu/}}  & Collection of parallel corpora in 500+ languages, gathered from many sources.   \\
WikiMatrix & {\footnotesize \url{https://bit.ly/3DrTjPo}} & Parallel corpora mined from Wikipedia \\
CCMatrix &{\small \url{https://bit.ly/3Bin6rQ}} &  Parallel corpora mined from CommonCrawl\\
Samanantar & {\footnotesize \url{https://indicnlp.ai4bharat.org/samanantar/}}  & A parallel corpus of 11 languages of India paired with English\\
Bible &{\footnotesize \url{https://github.com/christos-c/bible-corpus}} & A parallel corpus of 100 languages extracted from the Bible \\
Tanzil & {\footnotesize \url{https://opus.nlpl.eu/Tanzil.php}} &
A parallel corpus of 42 languages translated from the Quran by the Tanzil project.\\ 
\midrule
Tatoeba Challenge & {\small \url{https://bit.ly/3Drp36U}}& Test sets in over 100 language pairs. \\
WMT corpora &  {\small \url{http://www.statmt.org/wmt21/}} & Test (and training) sets for many shared tasks. \\
WAT corpora & {\footnotesize \url{https://lotus.kuee.kyoto-u.ac.jp/WAT/}} & Test (and training) sets for many shared tasks. \\
FLORES-101 & {\footnotesize \url{https://github.com/facebookresearch/flores}} & Test sets for 100 languages, paired with English\\
\midrule
Pavlick dictionaries & {\small \url{https://bit.ly/3DgI0cu}} & Crowd-sourced bilingual dictionaries in many languages \\ 
\bottomrule
\end{tabular}}
}
\caption{Summary of useful sources of monolingual, parallel and benchmarking data discussed in this section}
\label{tab:corpora}

\end{table}

\section{Use of monolingual data} 
\label{sec:monolingual}

For low-resource language pairs, parallel text is, by definition, in short supply, and even applying the data collection
strategies of Section~\ref{sec:data} may not yield sufficient text to train high-quality MT models.
However, monolingual text will almost always be more plentiful than parallel, and leveraging monolingual data has therefore been one of the most important and successful areas of research in low-resource MT. 

In this section, we provide an overview of the main approaches used to exploit monolingual data in low-resource scenarios. We start by reviewing work on integrating external language models into NMT (Section~\ref{sec:mono-external-lm}), work largely inspired from the use of language models in statistical MT (SMT). We then discuss research on synthesising parallel data (Section~\ref{sec:mono-via-synth}), looking at the dominant approach of backtranslation and its variants, unsupervised MT and the modification of existing parallel data using language models. Finally, we turn to transfer learning (Section~\ref{sec:mono-via-xfer}), whereby a model trained on monolingual data is used to initialise part or all of the NMT system, either through using pre-trained embeddings or through the initialising of model parameters from pre-trained language models.
The use of monolingual data for low-resource has previously been surveyed by \citet{gibadullin2019survey}, who choose to categorise methods according to whether they are architecture-independent or architecture-dependent. This categorisation approximately aligns with our split into (i)~approaches based on synthetic data and the integration of external LMs (architecture-independent), and (ii)~those based on transfer learning (architecture-dependent).\footnote{With the difference that unsupervised MT is architecture-dependent and we choose to discuss it in Section~\ref{sec:mono-via-synth} on synthesising parallel data.} 

\subsection{Integration of external language models}\label{sec:mono-external-lm}
For SMT, monolingual data was normally incorporated into the system using a language model, in a formulation that can be traced back to the noisy channel model \cite{brown-etal-1993-mathematics}. In early work on NMT, researchers drew inspiration from SMT, and several works have focused on integrating external language models into NMT models.

The first approach, proposed by \citet{GulcehreFXCBLBS15}, was to modify the scoring function of the MT model by either interpolating the probabilities from a language model with the translation probabilities (they call this \textit{shallow fusion}) or integrating the hidden states from the language model within the decoder (they call this \textit{deep fusion}). Importantly, they see improved scores for a range of scenarios, including a (simulated) low-resource language direction (Turkish$\rightarrow$English), with best results achieved using deep fusion. 
Building on this, \citet{stahlberg-etal-2018-simple} proposed \textit{simple fusion} as an alternative method for including a pre-trained LM. In this case, the NMT model is trained from scratch with the fixed LM, offering
it a chance to learn to be complementary to the LM. The result is improved translation performance as well
as training efficiency, with experiments again on low-resource Turkish--English, as well as on larger data sets for Xhosa$\rightarrow$English and Estonian$\rightarrow$English.

The addition of a language model to the scoring function as in the works described above has the disadvantage of increasing the time necessary for decoding (as well as training). An alternative approach was proposed by \citet{baziotis2020language}, who aim to overcome this by using the language model as a regulariser during training, pushing the source-conditional NMT probabilities to be closer to the unconditional LM prior.
They see considerable gains in very low-resource settings (albeit simulated), using small data sets for Turkish--English and German--English.

\subsection{Synthesising Parallel Data using Monolingual Data}
\label{sec:mono-via-synth}


One direction in which the use of monolingual data has been highly successful is in the production of synthetic parallel data. This is particularly important in low-resource settings when genuine parallel data is scarce. It has been the focus of a large body of research and has become the dominant approach to exploiting monolingual data due to the improvements it brings (particularly in the case of backtranslation as we shall see) and the potential for progress in the case of unsupervised MT. Both backtranslation and unsupervised MT belong to a category of approaches involving self-learning, which we discuss in Section~\ref{sec:monolingual_self-learning}. We then discuss an alternative method of synthesising parallel data in Section~\ref{sec:mono:lm-augmentation}, which involves modifying existing parallel data using a language model learnt on monolingual data.

\subsubsection{Self-learning: backtranslation and its variants}\label{sec:monolingual_self-learning}

One of the most successful strategies for leveraging monolingual data has been the creation of synthetic parallel data through translating monolingual texts either using a heuristic strategy or an intermediately trained MT model. This results in parallel data where one side is human generated, and the other is automatically produced. We focus here on backtranslation and its variants, before exploring in the next section how unsupervised MT can be seen as an extension of this idea.

\paragraph{Backtranslation}

Backtranslation corresponds to the scenario where target-side monolingual data is translated using an MT system to give corresponding synthetic source sentences, the idea being that it is particularly beneficial for the MT decoder to see well-formed sentences. Backtranslation was first introduced in SMT \citep{bertoldi-federico-2009-domain,bojar-tamchyna-2011-improving}, but since monolingual data
could already be incorporated easily into SMT systems using language models, and because
inference in SMT was quite slow, backtranslation was not widely used. For NMT however, it was discovered that backtranslation was a remarkably 
effective way of exploiting monolingual data \citep{sennrich2016improving}, and it remains an important technique, for both low-resource MT and MT in general. 

There has since been considerable interest in understanding and improving backtranslation. 
For example, \citet{edunov-etal-2018-understanding} showed that backtranslation improved performance even at a very large scale, but also that it provided improvements in 
(simulated) low-resource settings, where it was important to use beam-search rather than sampling to create backtranslations (the opposite situation to high-resource pairs). \citet{caswell-etal-2019-tagged} showed that simply adding a tag to the back-translated data during training, to let the model know which was back-translated data and which was natural data, could improve performance.

\paragraph{Variants of backtranslation}
Forward translation, where monolingual source data is translated into the target language \citep{zhang-zong-2016-exploiting}, is also possible but has received considerably less interest than backtranslation for low-resource MT, presumably because of the noise it introduces for the decoder. However, \citet{He2019-fr} actually find it more effective than backtranslation in their experiments for low-resource English$\rightarrow$Nepali when coupled with noising (as a kind of dropout), which they identify as an important factor in self-supervised learning. A related (and even simpler) technique related to forward and backtranslation, that of copying from target to source to create synthetic data, was introduced by \citet{currey2017copied}. They showed that this was particularly useful in low-resource settings (tested on Turkish--English and Romanian--English) and hypothesise that it helped particularly with the translation of named entities. 

\paragraph{Iterative backtranslation}
For low-resource NMT, back-translation can be a particularly effective way of improving quality \citep{guzman-etal-2019-flores}. However one possible issue is that the initial model used for  translation (trained on available parallel data) is often of poor quality when parallel data is scarce, which inevitable leads to poor quality backtranslations. The logical way to address this is to perform \emph{iterative backtranslation}, whereby intermediate models of increasing quality in both language directions are successively used to create synthetic parallel data for the next step. This has been successfully applied to low-resource settings by
several authors \citep{hoang2018iterative,dandapat2018eamt,bawden-etal-2019-university,sanchez-martinez-etal-2020-english}, although successive iterations offer diminishing returns, and often two iterations are sufficient, as has been shown experimentally \citep{chen-etal-2020-facebook}. 

Other authors have sought to improve on iterative backtranslation by introducing a round-trip (i.e.~autoencoder) objective for monolingual data, in other words performing backtranslation and forward translation implicitly during training. This was simultaneously proposed
by \citet{cheng2016semi} and \citet{NIPS2016_5b69b9cb} and also by \citet{zhang-zong-2016-exploiting} who also added forward translation. However, none of these authors applied their techniques to low-resource settings. In contrast, \citet{niu-etal-2019-bi} developed a method using Gumbel softmax to enable back-propagation through backtranslation: they tested in low-resource settings but achieved limited success.

Despite the large body of literature on applying back-translation and related techniques, and evidence that it works in low-resource
NMT, there are few systematic experimental study of back-translation specifically for low-resource NMT, apart from \citep{Xu2019-oe}, which appears to confirm the findings of \citep{edunov-etal-2018-understanding} that sampling is best when there are reasonable amounts of data and beam search is better when data is very scarce.

\subsubsection{Unsupervised MT}\label{sec:unsuper}

The goal of \emph{unsupervised MT} is to learn a translation model without any parallel data, so this can be considered an extreme form of low-resource
MT. The first unsupervised NMT models \citep{lample2018unsupervised,artetxe2018unsupervisednmt} were typically trained in a two-phase process: a rough translation system is first created by aligning word embeddings across the two languages (e.g.~using bilingual seed lexicons), and then several rounds of iterative backtranslation and denoising autoencoding are used to further train the system. Whilst this approach has been successfully applied to high-resource language pairs
(by ignoring available parallel data) it has been shown to perform poorly on genuine low-resource language pairs \citep{guzman-etal-2019-flores,marchisio-duh-koehn:2020:WMT,kim-etal-2020-unsupervised}, mainly because the initial quality of the word embeddings and their cross-lingual alignments is poor \citep{edman-etal-2020-low}. The situation is somewhat improved using transfer learning from models trained on large amounts of monolingual data (Section \ref{sec:mono-via-xfer}), and some further gains can be achieved by adding a supervised training step with the limited parallel data (i.e.~semi-supervised rather than unsupervised) \cite{bawden-etal-2019-university}. However the performance remains limited, especially compared to high-resource language pairs.

These negative results have focused researchers' attention on making unsupervised MT work better for low-resource languages. 
\citet{chronopoulou2021improving} improved the cross-lingual
alignment of word embeddings in order to get better results on unsupervised Macedonian--English and Albanian--English. A separate line of work is concerned with using corpora from other languages
to improve unsupervised NMT (see Section \ref{sec:multiunsuper})

\subsubsection{Modification of existing parallel data}\label{sec:mono:lm-augmentation}

Another way in which language models have been used to generate synthetic parallel data is to synthesise parallel examples from new ones by replacing certain words.\footnote{These techniques are inspired by data augmentation in computer vision, where it is much simpler to manipulate examples to create new ones (for example by flipping and cropping images) whilst preserving the example label. The difficulty in constructing synthetic examples in NLP in general is the fact that the modifying any of the discrete units of the sentence is likely to change the meaning or grammaticality of the sentence.} In translation, it is important to maintain the relation of translation between the two sentences in the parallel pair when modification of the pair occurs. There are to our knowledge few works so far in this area. \citet{fadaee2017data} explore data augmentation for MT for a simulated low-resource setting (using English--German). They rely on bi-LSTM language models to predict plausible but rare equivalents of words in sentences. They then substitute in the rare words and replace the aligned word in the corresponding parallel sentence with its translation (obtained through a look-up in an SMT phrase table). They see improved BLEU scores \citep{papineni-etal-2002-bleu} and find that it is a complementary technique to backtranslation. More recently, \citet{arthaud_fewshot_2021} apply a similar technique to improve the adaptation of a model to new vocabulary for the low-resource translation direction Gujarati$\rightarrow$English. They use a BERT language model to select training sentences that provide the appropriate context to substitute new and unseen words in order to create new synthetic parallel training sentences. While their work explores the trade-off between specialising to the new vocabulary and maintaining overall translation quality, they show that the approach can improve the translation of new words more effectively following data augmentation.

\subsection{Introducing Monolingual Data Using Transfer Learning}
\label{sec:mono-via-xfer}

The third category of approaches we explore looks at transfer learning, by which we refer to techniques where a model trained using the monolingual data is used to initialise some or all of the NMT model. A related, but different idea, multilingual models, where the low-resource NMT model may be trained with the help of other (high-resource) languages will be considered in Section~\ref{sec:multilingual}.

\paragraph{Pre-trained embeddings} When neural network methods were introduced to NLP, transfer learning meant using pre-trained word embeddings, such as word2vec \citep{2013arXiv1301.3781M} or GloVe \citep{pennington2014glove}, to introduce knowledge from large unlabelled monolingual corpora into the model. The later introduction of the multilingual fastText embeddings \citep{bojanowski-etal-2017-enriching} meant that pre-trained embeddings could be tested with NMT \citep{gangi2017iwslt,neishi-etal-2017-bag,qi-etal-2018-pre}. Pre-trained word embeddings were also used in the first phase of unsupervised NMT training (Section~\ref{sec:unsuper}). Of most interest for low-resource NMT was the study by \citet{qi-etal-2018-pre}, who showed that pre-trained embeddings could be extremely effective in some low-resource settings.

\paragraph{Pre-trained language models} Another early method for transfer learning was to pre-train a language model, and then to use it to initialise either the encoder or the decoder or both \citep{ramachandran-etal-2017-unsupervised}. Although not MT \textit{per se}, \citet {junczys-dowmunt-etal-2018-approaching} applied this method to improve grammatical error correction, which they modelled as a low-resource MT task.

The pre-trained language model approach has been extended with new objective functions based on predicting masked words, trained on large amounts of monolingual data. Models such as ELMo \citep{peters-etal-2018-deep} and BERT \citep{devlin2018bert} have been shown to be very beneficial to natural language understanding tasks, and researchers have sought to apply related ideas to NMT. One of the blocking factors identified by \citet{yang2020making} in using models such as BERT for pre-training is the problem of catastrophic forgetting \citep{Goodfellow2013-tt}. They propose a modification to the learning procedure involving a knowledge distillation strategy designed to retain the model's capacity to perform language modelling during translation. They achieve increased translation performance according to BLEU, although they do not test on low-resource languages. 

Despite the success of ELMo and BERT in NLP, large-scale pre-training in NMT did not become popular until the success of the 
 XLM \citep{lample2019cross}, MASS \citep{song2019mass} and mBART \citep{liu2020multilingual} models. These models allow transfer learning for
 NMT by initial training on large quantities of monolingual data in several languages, before fine-tuning on the languages of interest. Parallel data can also be incorporated into these pre-training approaches \cite{tang-etal-2021-multilingual}. Since they use data from several languages, they will be discussed
 in the context of multilingual models in Section~\ref{sec:multi-pretraining}.

\section{Use of multilingual data}
\label{sec:multilingual}

  
In the previous section we considered approaches that exploit monolingual corpora to compensate for the limited amount of parallel data available for low-resource language pairs. In this section we consider a different but related set of methods, which use additional data from different languages (i.e.\ in languages other than the language pair that we consider for translation). These multilingual approaches can be roughly divided into two
categories: (i)~transfer learning and (ii)~multilingual models. 

Transfer learning (Section~\ref{sec:multi-transfer}) was introduced in Section~\ref{sec:mono-via-xfer} in the context of pre-trained language models. 
These methods involve using some or all of the parameters of a ``parent'' model
to initialise the parameters of the ``child'' model.
The idea of multilingual modelling (Section~\ref{sec:multi-model})
is to train a system that is capable of translating between several different language 
pairs. This is relevant to low-resource MT, because low-resource language pairs included
in a multilingual model may benefit from other languages used to train the model. 
Finally (Section~\ref{sec:multi-pretraining}), we consider more recent approaches to transfer learning, based on learning large pre-trained models from multilingual collections of monolingual and parallel data.


\subsection{Transfer Learning}
\label{sec:multi-transfer}

In the earliest form of multilingual transfer learning for NMT, a parent model is trained
on one language pair, and then the trained parameters are used to initialise
a child model, which is then trained on the desired low-resource language pair.

This idea was first explored by \citet{zoph2016transfer}, who considered a French--English
parent model, and child models translating from 4 low-resource languages
(Hausa, Turkish, Uzbek and Urdu) into English. They showed that transfer learning could 
indeed improve over random initialisation, and the best performance for this scenario
was obtained when the values of the target embeddings were fixed after training
the parent, but the training continued for all the other parameters. 
\citet{zoph2016transfer} suggest that the choice of the parent language could be important, but did not explore this further for their low-resource languages.

Whilst \citet{zoph2016transfer} treat the parent and child vocabularies as independent, 
\citet{nguyen2017transfer} showed that when transferring between related languages
(in this case, within the Turkic family), it is beneficial to share
the vocabularies between the parent and child models. To boost this effect, subword
segmentation such as BPE \citep{sennrich2016neural} can help to further increase 
the vocabulary overlap. In cases where there is little vocabulary overlap 
(for example, because the languages are distantly related), mapping the bilingual
embeddings between parent and child can help \cite{kim2019effective}.
In cases where the languages are highly related but are written in different scripts, transliteration may be used to increase the overlap in terms of the surface forms \citep{dabre-etal-2018-nicts,goyal-etal-2020-efficient}. Interestingly, in the case of transferring from a high-resource language pair into a low-resource one where the target language is a variant of the initial parent language \citet{Kumar2021-ra} found it useful to pretrain embeddings externally and then fix them during the training of the parent, before initialising the embeddings of the low-resource language using those of the high-resource variant. Tested from English into Russian (transferring to Ukranian and Belarusian), Norwegian Bokmål (transferring to Nynorsk) and Arabic (transferring to four Arabic dialects), they hypothesise that decoupling the embeddings from training helps to avoid mismatch when transferring from the parent to the child target language.


The question of how to choose the parent language for transfer learning, as posed by \citet{zoph2016transfer}, has been taken up by later authors. One study suggests that language relatedness is important \citep{dabre-etal-2017-empirical}. However
\citet{kocmi2018trivial} showed that the main consideration in 
transfer learning is to have a strong parent model, and it can work well for
unrelated language pairs. Still, if languages are unrelated
and the scripts are different, for example transferring from an Arabic--Russian parent
to Estonian--English, transfer learning is less useful.  \citet{lin_choosing_2019} perform an extensive study on 
choosing the parent language for transfer learning, showing that data-related features
of the parent models and lexical overlap are often more important than language similarity.
Further insight into transfer learning for low-resource settings was provided by
\citet{aji-etal-2020-neural}, who analysed the training dynamics and concluded 
that the parent language is not important. The effectiveness from transfer learning 
with strong (but linguistically unrelated) parent models has been confirmed in shared task
submissions such as \citet{bawden-EtAl:2020:WMT1} -- see Section~\ref{sec:case-study:additional-data}.

Multi-stage transfer learning methods have also been explored. \citet{Dabre2019-nl} propose 
a two-step transfer with English on the source side for both parent and 
child models. First, a one-to-one parent model is used to initialise weights in a
multilingual one-to-many model, using a multi-way parallel corpora that includes
the child target language. Second, the intermediate multilingual model is fine-tuned on 
parallel data between English and the child target language. \citet{Kim2019-df} use a two-parent
model and a pivot language. One parent model is between the child source language and the pivot language (e.g.~German--English), and the other translates between the pivot and the child target language (e.g.~English--Czech). Then, the encoder parameters from the first model and the decoder parameters of the second models are used to initialise the parameters of the child model (e.g.~German--Czech).

\subsection{Multilingual Models}
\label{sec:multi-model}


The goal of multilingual MT is to have a universal model capable of translation between any 
two languages. Including low-resource language pairs in multilingual models can be seen as means
of exploiting additional data from other, possibly related, languages. Having more languages in 
the training data helps developing a universal representation space, which in turn allows for
some level of parameter sharing among the language-specific model components.

The degree to which parameters are shared across multiple language directions varies considerably in the literature, with early models showing little sharing across languages \citep{dong-etal-2015-multi} and some later models exploring the sharing of most or all parameters \citep{johnson2017google}. The amount of parameter sharing can be seen as a trade-off between ensuring that each language is sufficiently represented (has enough parameters allocated) and that low-resource languages can benefit from the joint training of parameters with other (higher-resource) language pairs (which also importantly reduces the complexity of the model by reducing the number of parameters required).


\citet{dong-etal-2015-multi} present one of the earliest studies in multilingual NMT, focused on translation from a single language into multiple languages simultaneously. The central idea of this approach is to have a shared encoder and many language-specific decoders, including language-specific weights in the attention modules. By training on multiple target languages (presenting as a multi-task setup), the motivation is that the representation of the source language will not only be trained on more data (thanks to the multiple language pairs), but the representation may be more universal, since it is being used to decode several languages. They find that the multi-decoder setup provides systematic gains over the bilingual counterparts, although the model was only tested in simulated low-resource settings.


As an extension to this method, 
\citet{firat-etal-2016-multi} experiment with multilingual models in the many-to-many scenario. They too use separate encoders and decoders for each language, but the attention mechanism is shared across all directions, which means adding languages increases the number of model parameters linearly (as opposed to quadratic increase when attention is language-direction-specific). In all cases, the multi-task model performed better than the bilingual models according to BLEU scores, although it was again only tested in simulated low-resource scenarios.

More recent work has looked into the benefits of sharing only certain parts of multilingual models, ensuring language-dependent components. For example, \citet{Platanios2018-bx} present a contextual parameter generator component, which allows finer control of the parameter sharing across different languages. \citet{fan_beyond_2020} also include language-specific components by sharing certain parameters across pre-defined language groups in order to efficiently and effectively upscale the number of languages included (see Section~\ref{sec:multi-massive}).

In a bid to both simplify the model (also reducing the number of parameters) and to maximally encourage sharing between languages, \citet{Ha2016-qo} and \citet{johnson2017google} proposed to use a single encoder and decoder to train all language directions (known as the universal encoder-decoder). Whereas \citet{Ha2016-qo} propose language-specific embeddings, \citet{johnson2017google} use a joint vocabulary over all languages included, which has the advantage of allowing shared lexical representations (and ultimately this second strategy is the one that has been retained by the community). The control over the target language was ensured in both cases by including pseudo-tokens indicating the target language in the source sentence. Although not trained or evaluated on low-resource language pairs, the model by \citet{johnson2017google} showed promise in terms of the ability to model multilingual translation with a universal model, and zero-shot translation (between language directions for which no parallel training data was provided) was also shown to be possible. The model was later shown to bring improvements when dealing with translation into several low-resource language varieties \citep{Lakew2018-or}, a particular type of multilingual MT where the several target languages are very similar. 
We shall see in the next section (Section~\ref{sec:multi-massive}) how scaling up the number of languages used for training can be beneficial in the low-resource setting.

Combining multilingual models, with the transfer learning approaches of the previous section,
\citet{neubig2018rapid} present a 
number of approaches for adaptation of multilingual models to new languages.
The authors consider cold- and warm-start scenarios, depending on whether the
training data for the new language was available for training the original
multilingual model. They find that multilingual models fine-tuned with the 
low-resource language training data mixed in with data from a similar
high-resource language (i.e. similar-language regularisation) give the best 
translation performance.

\subsubsection{Massively multilingual models}\label{sec:multi-massive}
In the last couple of years, efforts have been put into scaling up the number of languages included in multilingual training, particularly for the universal multilingual model \citep{johnson2017google}. The motivation is that increasing the number of languages should improve the performance for all language directions, thanks to the addition of extra data and to increased transfer between languages, particularly for low-resource language pairs.
For example, \citet{neubig2018rapid} trained a many-to-English model with 57 possible source languages, and more recent models have sought to include even more languages; \citet{aharoni2019massively} train an MT model for 102 languages to and from English as well as a many-to-many MT model between 59 languages, and  \citet{fan_beyond_2020}, \citet{zhang2020improving} and \citet{Arivazhagan2019-iv} train many-to-many models for over 100 languages.

While an impressive feat, the results show that it is non-trivial to maintain high translation performance across all languages as the number of language pairs is increased \citep{mueller2020analysis,aharoni2019massively,Arivazhagan2019-iv}. There is a trade-off between \textit{transfer} (how much benefit is gained from the addition of other languages) and \textit{interference} (how much performance is degraded due to having to also learn to translate other languages) \citep{Arivazhagan2019-iv}. It is generally bad news for high-resource language pairs, for which the performance of multilingual models is usually below that of language-direction-specific bilingual models. However, low-resource languages often do benefit from multilinguality, and the benefits are more
noticeable for the many-to-English than for the English-to-many \cite{johnson2017google,Arivazhagan2019-iv}. It has also been shown that for zero-shot translation, the more languages included in the training, the better the results are \citep{aharoni2019massively,Arivazhagan2019-iv}, and that having multiple bridge pairs in the parallel data (i.e.~not a single language such as English being the only common language between the different language pairs) greatly benefits zero-shot translation, even if the amount of non-English parallel data remains small \citep{Rios2020-qb,Freitag2020-qp}.

There is often a huge imbalance in the amount of training data available across language pairs and for low-resource language pairs, it is beneficial to upsample the amount of data. However, upsampling low-resource pairs has the unfortunate effect of harming performance on high-resource pairs \citep{Arivazhagan2019-iv}, and there is the additional issue of the model overfitting on the low-resource data even before it has time to converge on the high-resource language data. A solution to this problem is the commonly used strategy of temperature-based sampling, which involves adjusting how much we sample from the true data distribution \citep{devlin2018bert,fan_beyond_2020}, providing a certain compromise between making sure low-resource languages are sufficiently represented but reducing the deterioration in performance seen in high-resource language pairs. Temperature-based sampling can also be used when training the subword segmentation to create a joint vocabulary across all languages so that the low-resource languages are sufficiently represented in the joint vocabulary despite there being little data.

Several works have suggested that the limiting factor is the capacity of the model (i.e.~the number of parameters). Whilst multilingual training with shared parameters can increase transfer, increasing the number of languages decreases the per-task capacity of the model. \citet{Arivazhagan2019-iv} suggest that model capacity may be the most important factor in the transfer-interference trade-off; they show that larger models (deeper or wider) show better translation performance across the board, deeper models being particularly successful for low-resource languages, whereas wider models appeared more prone to overfitting.
\citet{zhang2020improving} show that online backtranslation combined with a deeper Transformer architecture and a special language-aware layer normalisation and linear transformations between the encoder and the decoder improve translation in a many-to-many setup.


\subsubsection{Multilingual Unsupervised Models}
\label{sec:multiunsuper}
As noted in Section~\ref{sec:unsuper}, unsupervised MT performs quite poorly in low-resource language pairs, and one of the ways
in which researchers have tried to improve its performance is by exploiting data from other languages. \citet{sen_multilingual_2019} demonstrate that a multilingual unsupervised NMT model can perform better than bilingual models in each language pair,
but they only experiment with high-resource language pairs. Later works \citep{garcia_harnessing_2020,ko-etal-2021-adapting} directly address the problem of 
unsupervised NMT for a low-resource language pair in the case where there is parallel data in a related language. More
specifically, they use data from a third language ($Z$) to improve unsupervised MT between a low-resource language ($X$) and a high-resource language ($Y$). In both works, they assume
that $X$ is closely related to $Z$, and that there is parallel data between $Y$ and $Z$. As in the original unsupervised NMT
models \citep{lample2018unsupervised,artetxe2018unsupervisednmt}, the training process uses denoising autoencoders and iterative backtranslation.

\subsection{Large-scale Multilingual pre-training}\label{sec:multi-pretraining}
The  success of large-scale pre-trained language models such as ELMo \citep{peters-etal-2018-deep} and BERT \citep{devlin2018bert} has inspired researchers to apply related techniques to MT. Cross-lingual language models (XLM; \citealp{lample2019cross}) are a direct application of the BERT masked language model (MLM) objective to learn from parallel data. The training data consists of concatenated sentence pairs, so that the model learns to predict the identity of the masked words from the context in both languages simultaneously. XLM was not applied to low-resource MT in the original paper, but was shown to improve unsupervised MT, as well as language modelling and natural language inference in low-resource languages. 

The first really successful large-scale pre-trained models for MT were mBART \citep{liu2020multilingual} and MASS \citep{song2019mass}, which demonstrated improvements to NMT in supervised, unsupervised and semi-supervised (i.e.~with back-translation) conditions, including low-resource language pairs. The idea of these models is to train a noisy autoencoder using large collections of monolingual (i.e.\ not parallel) data in 2 or more languages. The autoencoder is a transformer-based encoder-decoder, and the noise is introduced by randomly masking portions of the input sentence. Once the autoencoder has been trained to convergence, its parameters can be used to initialise the MT model, which is trained as normal. Using mBART, \citet{liu2020multilingual} were able to demonstrate unsupervised NMT working on the distant low-resource language pairs Nepali--English and Sinhala--English, as well as showing improvements in supervised NMT in low-resource language pairs such as Gujarati--English.

The original mBART was trained on 25 different languages and its inclusion in HuggingFace \citep{wolf-etal-2020-transformers} makes it straightforward to use for pre-training. It has since been extended to mBART50 \cite{tang-etal-2021-multilingual}, which is trained on a mixture of parallel and monolingual data, and includes 50 different languages (as the name suggests); mBART50 is also available on HuggingFace. A recent case study \citep{birch-etal-2021-surprise} has demonstrated that mBART50 can be combined with focused data gathering techniques to quickly develop a domain-specific, state-of-the-art MT system for a low-resource language pair (in this case, Pashto--English).

A recent multilingual pre-trained method called mRASP~\cite{lin2020pre}  has shown strong performance across a range of MT tasks: medium, low and very low-resource. mRASP uses unsupervised word alignments generated by MUSE~\cite{Conneau2018-uj} to perform random substitutions of words with their translations in another language, with the aim of bringing words with similar meanings across multiple languages closer in the representation space. They show gains of up to 30 BLEU points for some very low-resource language pairs such as Belarusian--English. mRASP2~\cite{pan2021contrastive} extends this work by incorporating monolingual data into the training. 

Of course, pre-trained models are useful if the languages you are interested in are included in the pre-trained model, and you have the resources to train and deploy these very large models. On the former point, \citet{muller-etal-2021-unseen} have considered the problem of extending multilingual BERT (mBERT) to new languages for natural language understanding tasks. They find greater difficulties for languages which are more distant from those in mBERT and/or have different scripts -- but the latter problem can be mitigated with careful transliteration.


\section{Use of external resources and linguistic information}
\label{sec:linguistic}

For some languages, alternative sources of linguistic information, for example (i)~linguistics tools (Section~\ref{sec:ling:tools-resources}) and (ii)~bilingual lexicons (Section~\ref{sec:ling:lexicons}), can be exploited. They can provide richer information about the source or target languages (in the case of tagging and syntactic analysis) and additional vocabulary that may not be present in parallel data (in the case of bilingual lexicons and terminologies).
Although there has been a large body of work in this area in MT in general, only some have been applied to true low-resource settings. We assume that this is because of  the lack of tools and resources for many of these languages, or at least the lack of 
those of sufficiently good quality. We therefore review work looking at exploiting these two sources of additional information, for languages where such resources are available and put a particular focus on those that have been applied to low-resource languages.

\subsection{Linguistic tools and resources}\label{sec:ling:tools-resources}

Additional linguistic analysis such as part-of-speech tagging, lemmatisation 
and parsing can help to reduce sparsity by providing abstraction from surface forms, as long as the linguistic tools and resources are available. A number of different approaches have developed for the integration of linguistic information in NMT. These include morphological segmentation, factored representations (Section~\ref{sec:linguistic_factored}), multi-task learning (Section~\ref{sec:linguistic_multitask}), interleaving of annotations (Section~\ref{sec:linguistic_interleaving}) and syntactic reordering (Section~\ref{sec:linguistic_reordering}).
At the extreme, these resources can be used to build full rule-based translation models (Section~\ref{sec:ml:rulebased}).

\subsubsection{Morphological segmentation}
A crucial part of training NMT system is the choice of subword segmentation, a pre-processing technique providing the ability to represent an infinite vocabulary with a fixed number of units and to better generalise over shorter units. For low-resource languages, it is even more important because there is a greater chance of coming across words that were not seen at training time. The most commonly used strategies are statistics-based, such as BPE \citep{sennrich2016neural} and sentencepiece \citep{kudo-richardson-2018-sentencepiece}. Not only might these strategies not be optimal from a point of view of linguistic generalisation, but for low-resource languages especially they have also been shown to give highly variable results, depending on what degree of segmentation is selected; this degree is a parameter which therefore must be chosen wisely \cite{ding-etal-2019-call,sennrich2019revisiting}.

Works exploring linguistic subword segmentation go back to statistical MT \cite{Oflazer2007-fp,Goldwater2005-xy}. Much of the focus has been on morphologically rich languages, with high degrees of inflection and/or compounding, for example for German, where minor gains can be seen over standard BPE \citep{Huck2017-ro}.
Specifically for low-resource languages, several works have tested the use of morphological analysers to assist the segmentation of texts into more meaningful units. In their submission to the WMT19 shared task for English$\rightarrow$Kazakh, \citet{sanchez-cartagena-etal-2019-universitat} use the morphological analyser from Apertium \citep{forcada-tyers-2016-apertium} to segment Kazakh words into stem (often corresponding to the lemma in Kazakh) and the remainder of the word. They then learnt BPE over the morphological segmented data.
\citet{Ortega2020-ap} also use a BPE approach, guided by a list of suffixes, which are provided to the algorithm and are not segmented. They see better performance than using Morfessor or standard BPE.
\citet{saleva-lignos-2021-effectiveness} also test morphologically aware subword segmentation for three low-resource language pairs: Nepali, Sinhala and Kazakh to and from English. They test segmentations using the LMVR \cite{Ataman2017-sb} and MORSEL \citep{lignos_learning_2010} analysers, but found no gain over BPE and no consistent pattern in the results. These results go against previous results from \citet{gronroos-etal-2014-morfessor} that showed that an LMVR segmentation can outperform BPE when handling low-resource Turkish, but they are in accordance with more recent ones for Kazakh--English \citep{toral-EtAl:2019:WMT} and Tamil--English \citep{dhar-etal-2020-linguistically}, where it does not seem to improve over BPE.

\subsubsection{Factored models}\label{sec:linguistic_factored}
Factored source and target representations \citep{garciamartinez:hal-01433161,sennrich-haddow-2016-linguistic,burlot-etal-2017-word} were designed as a way of decomposing word units into component parts, which can help to provide some level of composite abstraction from the original wordform. For example, a wordform may be represented by its lemma and its part-of-speech, which together can be used to recover the original surface form. This type of modelling can be particularly useful for morphologically rich languages (many of which are already low-resource), for which the large number of surface forms can result in greater data sparsity and normally necessitate greater quantities of data.


Factored models originated in SMT \citep{koehn-hoang-2007-factored}, but were notably not easy to scale. The advantage of factored representations in NMT is that the factors are represented in continuous space and therefore may be combined more easily, without resulting in an explosion in the number of calculations necessary. 
\citet{garciamartinez:hal-01433161},  \citet{sennrich-haddow-2016-linguistic} and \citet{burlot-etal-2017-word} evaluate on language pairs involving at least one morphologically rich language and show that improvements in translation quality can be seen, but this is dependent on the language pair and the type of linguistic information included in the factors. \citet{nadejde-etal-2017-predicting} use factors to integrate source-side syntactic information in the form of CCG tags \citep{Steedman2000Syntactic,Clark2007CCG}, which they combine with an interleaving approach on the target side (see Section~\ref{sec:linguistic_interleaving}) to significantly improve MT performance, for high-resource (German$\rightarrow$English) and mid-low-resource (Romanian$\rightarrow$English) language directions.

\subsubsection{Multi-task learning}\label{sec:linguistic_multitask}
Multi-task learning can be seen as a way of forcing the model to learn better internal representations of wordforms by training the model to produce a secondary output (in this case linguistic analyses) as well as a translation.

Initial work in multi-task learning for MT did not concentrate on the low-resource scenario.
\citet{luong2016multitask} explore different multi-task setups for translation (testing on English--German), among which a setup where they use parsing as an auxiliary task to translation, which appears to help translation performance as long as the model is not overly trained on the parsing task. The question of how to optimally train such multi-task models has inevitably since been explored, inspired in part by concurrent work in multi-encoder and multi-decoder multilingual NMT (See Section~\ref{sec:multilingual}), since it appears that sharing all components across all tasks is not the optimal setting. \citet{niehues-cho-2017-exploiting} experiment with part-of-speech (PoS) tagging and named entity recognition as auxiliary tasks to MT and test different degrees of sharing. They find that sharing the encoder only (i.e.~separate attention mechanisms and decoders) works best and that using both auxiliary tasks enhances translation performance in a simulated low-resource German$\rightarrow$English scenario.

Since then, there have been some applications of multi-task learning to lower-resource scenarios, with slight gains in translation performance. \citet{nadejde-etal-2017-predicting} also share encoders in their multi-task setting for the integration of target-side syntactic information in the form of CCG supertags (for German$\rightarrow$English and mid-low-resource Romanian$\rightarrow$English). Similarly, \citet{zaremoodi2018adaptive} develop a strategy to avoid task interference in a multi-task MT setup (with named entity recognition, semantic parsing and syntactic parsing as auxiliary tasks). They do so by extending the recurrent components of the model with multiple blocks and soft routing between them to act like experts. They test in real low-resource scenarios (Farsi--English and Vietnamese--English) and get gains of approximately 1 BLEU point by using the additional linguistic information in the dynamic sharing setup they propose.



\subsubsection{Interleaving of linguistic information in the input}\label{sec:linguistic_interleaving}


As well as comparing factored representations and multi-task decoding, \citet{nadejde-etal-2017-predicting} also introduce a new way of integrating target-side syntactic information, which they call \textit{interleaving}. The idea is to annotate the target side of the training data with token-level information (CCG supertags in their case) by adding a separate token before each token containing the information pertaining to it, so that the model learns to produce the annotations along with the translation. They found this to work better than multi-task for the integration of target-side annotations and was also complementary with the use of source factors.
Inspired by these results, \citet{tamchyna-etal-2017-modeling} also followed the interleaving approach (for English$\rightarrow$Czech and English$\rightarrow$German, so not low-resource scenarios), but with the prediction of interleaved morphological tags and lemmas, followed by a deterministic wordform generator. Whereas \citet{nadejde-etal-2017-predicting} seek to create representations that are better syntactically informed, the aim of \citep{tamchyna-etal-2017-modeling} is different: they aim to create a better generalisation capacity for translation into a morphologically rich language by decomposing wordforms into their corresponding tags and lemmas. They see significantly improved results with the two-step approach, but find that simply interleaving morphological tags (similar to \citet{nadejde-etal-2017-predicting}) does not lead to improvements. They hypothesise that the morphological tags are less informative than CCG supertags and therefore the potential gain in information is counterbalanced by the added difficulty of having to translate longer target sequences.

In a systematic comparison with both RNN and transformer architectures and for 8 language directions (and in particular for low-resource languages), \citet{sanchez-understanding-2020} find that interleaving (with part-of-speech information and morphological tags) is beneficial, in line with the conclusions from \citet{nadejde-etal-2017-predicting}. Interestingly, they find that (i)~interleaving linguistic information in the source sentence can help, and morphological information is better than PoS tags and (ii)~interleaving in the target sentence can also help, but PoS tagging is more effective than morphological information, despite the translations being more grammatical with added morphological information.

\subsubsection{Syntactic Reordering}\label{sec:linguistic_reordering}
Other than being used as an additional form of input, syntactic information can also be used \textit{a priori} to facilitate the translation task by reordering words within sentences to better match a desired syntactic order. 
\citet{Murthy2019-gj} found this to be particularly effective for very low-resource languages in a transfer learning setup, when transferring from a high-resource language pair to a low-resource pair (see Section~\ref{sec:multi-transfer}). 
Testing on translation into Hindi from Bengali, Gujarati, Marathi, Malayalam and Tamil, having transferred from the parent language direction English$\rightarrow$Hindi, they apply syntactic reordering rules on the source-side to match the syntactic order of the child source language, resulting in significant gains in translation quality.

\subsection{Bilingual lexicons}\label{sec:ling:lexicons}

Bilingual lexicons are lists of terms (words or phrases) in one language associated with their translations in a second language. The advantage of bilingual lexicons is that they may well provide specialist or infrequent terms that do not appear in available parallel data, with the downside that they do not give information about the translation of terms in context, notably when there are several possible translations of a same term. However, they may be important resources to exploit, since they provide complementary information to parallel data and may be more readily available and cheaper to produce.\footnote{Note that many of the works designed to incorporate bilingual lexicons actually work on automatically produced correspondences in the form of phrase tables (produced using SMT methods). Although these may be extracted from the same parallel data as used to train the NMT model, it may be possible to give more weight to infrequent words than may be the case when the pairs are learnt using NMT.}

The approaches developed so far to exploit bilingual lexicons to directly improve NMT can be summarised as follows: (i)~as seed lexicons to initialise unsupervised MT \citep{lample-etal-2018-phrase,duan-etal-2020-bilingual} (as described in Section~\ref{sec:unsuper}), (ii)~as an additional scoring component, particularly to provide coverage for rare of otherwise unseen vocabulary \citep{arthur-etal-2016-incorporating,Feng-etal-2017-memory} and (iii)~as annotations in the source sentence by adding translations from lexicons just after their corresponding source words \cite{dinu-etal-2019-training}\footnote{The origin of the source units (i.e.~original or inserted translation) is distinguished by using factored representations. A similar technique was used to insert dictionary definitions rather than translations by \citet{Zhong2020-pa}.} or by replacing them in a code-switching-style setup \citep{Song2019-xl}.

The most recent work on using lexicons in pretrained multilingual models \citep[mRASP]{lin2020pre} shows the most promise. Here translations of words are substituted into the source sentence in pretraining, with the goal of bringing words with similar meanings across multiple languages closer
in the representation space. 
Please see Section~\ref{sec:multi-pretraining} for more details. 


\section{Model-centric Techniques} 
\label{sec:ml}

In the previous sections we have looked at using monolingual data, data from other language pairs and other linguistic data to improve translation. In this section we explore work that aims to make better use of the data we already have by investigating better modelling, training and inference techniques.

In recent years, MT systems have converged towards a fairly standardised architecture: a sequence-to-sequence neural network model with an encoder and an auto-regressive decoder, typically implemented as a Transformer \cite{vaswani2017attention}, although recurrent models \citep{Bahdanau2015seqtoseq} are still used.
Training is performed on a parallel corpus by minimising the cross-entropy of the target translations conditional on the source sentences.
Monolingual examples, if available, are typically converted to parallel sentences, as discussed in Section~\ref{sec:monolingual}.
Once the model is trained, translations are usually generated by beam search with heuristic length control, which return high-probability sentences according to the learned distribution of target sentences conditional on the source sentences.
 
This approach has been very successful for MT on high-resource language pairs where there is enough high-quality parallel and monolingual text covering a wide variety of domains to wash out most of the misaligned inductive bias\footnote{Inductive bias is the preference towards certain probability distributions that the system has before training, resulting from its architecture.} that the model might have.
However, for low-resource language pairs the inductive bias of the model becomes more prominent, especially when the model operates out of the training distribution, as it frequently does when the training data has sparse coverage of the language. 
Therefore it can be beneficial to design the neural network architecture and training and inference procedures to be more robust to low-resource conditions, for instance by explicitly modelling the \textit{aleatoric} uncertainty\footnote{Uncertainty that is caused by the intrinsic randomness of the task, as opposed to \textit{epistemic} uncertainty which results from ignorance about the nature of the task.} that is intrinsic to the translation task due to its nature of being a many-to-many mapping (one source sentence can have multiple correct translations and one translation can result from multiple source sentences \citep{ott2018analyzing}).

In this section, we will review recent machine learning techniques that can improve low-resource MT, such as meta-learning for data-efficient domain adaptation  and multilingual learning (Section~\ref{sec:ml:meta}), 
Bayesian and latent variable models for explicit quantification of uncertainty (Section~\ref{sec:ml:latent}),  
alternatives to cross-entropy training (Section~\ref{sec:ml:train}) 
and beam search inference (Section~\ref{sec:ml:inference}). 
We will also briefly discuss rule-based approaches for translation between related low-resource languages (Section~\ref{sec:ml:rulebased}).

\subsection{Meta Learning}
\label{sec:ml:meta}
In Section~\ref{sec:multilingual} we discussed using multilingual training to improve low-resource MT by combining training sets for different language pairs in joint-learning or transfer learning schemes.
A more extreme form of this approach involves the application of \emph{meta learning}: rather than training a system to directly perform well on a single task or fixed set of tasks (language pairs in our case), a system can be trained to quickly adapt to a novel task using only a small number of training examples, as long as this task is sufficiently similar to tasks seen during (meta-)training.

One of the most successful meta-learning approaches is Model-Agnostic Meta-Learning (\emph{MAML}) \citep{pmlr-v70-finn17a}, which was applied to multilingual MT by \citet{gu2018meta}.
In MAML, we train task-agnostic model parameters $\bar{\theta}$ so that they can serve as a good initial value that can be further optimised towards a task-specific parameter vector $\overset{*}{\theta}_{m}$ based on a task-specific training set $D_{m}$.
This is accomplished by repeatedly simulating the fine-tuning procedure, evaluating each fine-tuned model on its task-specific evaluation set and then updating the task-agnostic parameters in the direction that improves this score.

Once training is completed, the fine-tuning procedure can be directly applied to any novel task.
\citet{gu2018meta} apply MAML by meta-training on synthetic low-resource tasks obtained by randomly subsampling parallel corpora of high-resource language pairs and then fine-tune on true low-resource language pairs, obtaining substantial improvements.

An alternative approach to meta-learning involves training \emph{memory-augmented networks} that receive the task-specific training examples at execution time and maintain a representation of them which they use to adapt themselves on the fly \citep{Vinyals2016,Santoro2016}, an approach related to the concept of ``fast weights'' computed at execution time as opposed to ``slow weights'' (the model parameters) computed at training time \citep{Schmidhuber1992}.
\citet{Lake2019} applied memory-augmented networks to synthetic sequence-to-sequence tasks in order to evaluate out-of-distribution generalisation under a variety of conditions.
Curiously, very large language models such as GPT-2 \citep{radford2019language} and in particular GPT-3 \citep{Brown2020} also exhibit this meta-learning capability even without any modification of the network architecture or training procedure, suggesting that meta-learning itself can be learned from a sufficient large amount of data.
In fact, GPT-3 achieves near-SOTA quality when translating into English even with a single translation example, for multiple source languages including Romanian, a medium-low resource language.

\subsection{Latent variable models}
\label{sec:ml:latent}

Auto-regressive NMT models 
can in principle represent arbitrary probability distributions given enough model capacity and training data. However in low-resource conditions, the inductive biases of the models might be insufficient for a good generalisation, and different factorizations of probability distributions that result in different inductive biases may be beneficial.

Various approaches have attempted to tackle these issues by introducing \emph{latent variables}, random variables that are neither observed as source sentences nor as target sentences, and are rather inferred internally by the model.
This can be done with a \emph{source-conditional} parametrisation, which applies latent-variable modelling only on the target sentence
or with a \emph{joint} parametrisation, which applies it to both the source and the target sentences.

Latent variable models enable a higher model expressivity and more freedom in the engineering of the inductive biases, at the cost of  more complicated and computationally expensive training and inference.
For this reason, approximation techniques such as Monte Carlo sampling or MAP inference over the latent variable are used, typically based on the \emph{Variational Autoencoder} framework (VAE) \citep{Kingma2014, pmlr-v32-rezende14}.

In the earliest Variational NMT approach by \citet{zhang-etal-2016-variational-neural}, a source-conditional parametrisation is used 
and the latent variable is a fixed-dimension continuous variable that is intended to capture global information about the target sentence.
Training is performed by maximising a lower bound, known as the \emph{Evidence Lower Bound} (ELBO) of the conditional cross-entropy of the training examples, 
which is computed using an auxiliary model component known as \emph{inference network} 
that approximates the posterior of the latent variable as a diagonal Gaussian conditional on both the source and the target sentence. 
During inference the latent variable is either sampled from the prior or more commonly approximated as its mode (which is also its mean).
This basic approach, similar to image VAEs and the Variational Language Model of \citet{bowman-etal-2016-generating}, adds limited expressivity to autoregressive NMT because a fixed-dimensional unimodal distribution is not especially well suited to represent the variability of a sentence, but it can be extended in various ways: \citet{su2018varrecnmt} and \citet{schulz-etal-2018-stochastic} use a sequence of latent variables, 
one for each target token, parametrised with temporal dependencies between each other.

\citet{setiawan-etal-2020-variational} parametrise the latent posterior 
using \emph{normalising flows} \citep{pmlr-v37-rezende15}, which can represent arbitrary and possibly multi-modal distributions as a sequence of transformation layers applied to a simple base distribution.

\citet{Eikema2019} use a joint parametrisation 
as they claim that explicitly modelling the source sentence together with the target sentence provides additional information to the model.
Inference is complicated by the need to post-condition the joint probability distribution on the source sentence, hence a series of approximations is used in order to ensure efficiency.

The latent variable models described so far have been evaluated on high-resource language pairs, although most of them have been evaluated on the IWSLT dataset, which represents a low-resource domain.
However, latent-variable MT has also been applied to fully low-resource language pairs, using models where the latent variables have been designed to have linguistically-motivated inductive bias.
\citet{ataman2019latent} introduce a NMT model with latent word morphology in a hierarchical model, allowing for both word level representations and character level generation to be modelled. 
This is beneficial for morphologically rich languages, which include many Turkic and African low-resource languages.
These languages use their complex morphologies to express syntactic and semantic nuance, which might not be captured by the purely unsupervised and greedy BPE preprocessing, especially when the BPE vocabulary is trained on a small corpus.
The proposed model uses for each word one multivariate Gaussian latent variable representing a lemma embedding and a sequence of quasi-discrete latent variables representing morphology.
Training is performed in a variational setting using a relaxation based on the Kumaraswamy distribution \citep{KUMARASWAMY198079, louizos2018learning, bastings-etal-2019-interpretable}, and inference is performed by taking the modes of the latent distributions, as the model is source-conditional.
This approach has been evaluated on morphologically rich languages including Turkish, yielding improvements both in in-domain and out-of-domain settings.


\subsection{Alternative training objectives}
\label{sec:ml:train}

When an autoregressive model is trained to optimise the cross-entropy loss, it is only exposed to ground-truth examples during training.
When this model is then used to generate a sequence with ancestral sampling, beam search or another inference method, it has to incrementally extend a prefix that it has generated itself.
Since the model in general cannot learn exactly the ``true'' probability distribution of the target text, the target prefix that it receives as input will be out-of-distribution, which can cause the estimate of the next token probability to become even less accurate. This issue, named \emph{exposure bias} by \citet{ranzato2016sequence}, can compound with each additional token and might result in the generated text to eventually become completely nonsensical.
Exposure bias theoretically occurs regardless of the task, but while its impact has been argued to be small in high-resource settings \citep{wu-etal-2018-beyond}, in low-resource MT it has been shown to be connected to the phenomenon of \emph{hallucination}, where the system generates translations that are partially fluent but contain spurious information not present in the source sentence \citep{wang2020exposure}.

A number of alternatives to cross-entropy training have been proposed in order to avoid exposure bias, which all involve exposing the model during training to complete or partial target sequences generated by itself.
\citet{ranzato2016sequence} explore multiple training strategies and propose a method called \emph{MIXER} which is a variation of the  \emph{REINFORCE} algorithm \citep{Williams1992, DBLP:journals/corr/ZarembaS15}.
In practice \mbox{REINFORCE} suffers from high variance, therefore they apply it only after the model has already been pre-trained with cross-entropy, a technique also used by all the other training methods described in this section.
They further extend the algorithm by combining cross-entropy training and REINFORCE within a each sentence according to a training schedule which interpolates from full cross-entropy to full REINFORCE.
They do not evaluate on a true low-resource language pair, but they do report improvements on German$\rightarrow$English translation on the relatively small IWSLT 2014 dataset \citep{cettolo2014report}.

\emph{Contrastive Minimum Risk Training} (CMRT or just MRT) \citep{och-2003-minimum, shen-etal-2016-minimum, edunov-etal-2018-classical} is a similar training technique that can be considered a biased variant of REINFORCE that focuses on high-probability translations generated by decoding from the model itself.
\citet{wang2020exposure} apply CMRT to low-resource translation (German$\rightarrow$Romansh) as well as German$\rightarrow$English IWSLT 2014, reporting improvements in the out-of-domain test case, as well as a reduction of hallucinations.

Both REINFORCE and CMRT use a reward function that measures the similarity between generated and reference translations, often based on an automatic evaluation metric such as BLEU.
However, the exact mechanism that makes such approaches work is not completely clear, \citet{Choshen2020On} show that REINFORCE and CMRT also work when the reward function is a trivial constant function rather than a sentence similarity metric, suggesting that their primary effect is to regularise the model pretrained with cross-entropy by exposing it to its own translations, hence reducing exposure bias.

An alternative training technique involving beam search decoding in the training loop has been proposed by \citet{wiseman-rush-2016-sequence}, based on the \emph{LaSO} \citep{Daume2005} structured learning algorithm.
This approach also exposes the model to its own generations during training, and it has the benefit that training closely matches the inference process, reducing any mismatch. The authors report improvements on German$\rightarrow$English IWSLT 2014. However they do not evaluate on a true low-resource language pair.

An even simpler technique that exposes the model's own generations during training is \emph{scheduled sampling} \citep{Bengio2015}, which also starts with cross-entropy training and progressively replaces part of the ground truth target prefixes observed by the model with its own samples.
Plain scheduled sampling is theoretically unsound \citep{DBLP:journals/corr/Huszar15}, but it can be made more consistent by backpropagating gradients through a continuous relaxation of the sampling operation, as shown by \mbox{\citet{xu-etal-2019-differentiable}} who report improvements on the low-resource Vietnamese$\rightarrow$English language pair.

Regularisation techniques have also been applied to low-resource MT.
\citet{sennrich2019revisiting} evaluated different hyperparameter settings, in particular batch size and dropout regularisation, for German$\rightarrow$English with varying amounts of training data and low-resource Korean$\rightarrow$English.
\citet{muller2019domain} experimented with various training and inference techniques for out-of-distribution MT both for a high-resource (German$\rightarrow$English) and low-resource (German$\rightarrow$Romansh) pair.
For the low-resource pair they report improvements by using sub-word regularisation \citep{kudo-2018-subword}, defensive distillation and source reconstruction.
An alternate form of subword regularisation, known as BPE dropout has been proposed by \citet{Provilkov2019bpedropout}, reporting improvements on various high-resource and low-resource language pairs.
\citet{he-etal-2020-dynamic} apply a dynamic programming approach to BPE subword tokenisation, evaluating during training all possible ways of tokenising each target word into subwords, and computing an optimal tokenisation at inference time.
Since their method is quite slow however, they only use it to tokenise the training set and then train a regular Transformer model on it, combining it with BPE dropout on source words, reporting improvements on high-resource and medium-resource language pairs.

\subsection{Alternative inference algorithms}
\label{sec:ml:inference}

In NMT, inference is typically performed using a type of beam search algorithm with heuristic length normalisation\footnote{The implementations of beam search differ between MT toolkits in details that can have significant impact over translation quality and are unfortunately often not well documented in the accompanying papers.} \citep{jean-etal-2015-montreal, koehn2017six}.
Ostensibly, beam search seeks to approximate maximum a posteriori (MAP) inference, however it has been noticed that increasing the beam size, which improves the accuracy of the approximation, often degrades translation quality after a certain point \citep{koehn2017six}.
It is actually feasible to exactly solve the maximum a posteriori inference problem, and the resulting mode is often an abnormal sentence; in fact, it is often the empty sentence \cite{stahlberg2019nmt}.
It is arguably dismaying that NMT relies on unprincipled inference errors in order to generate accurate translations.
Various authors have attributed this ``beam search paradox'' to modelling errors caused by exposure bias or other training issues and they have proposed alternative training schemes such as these discussed in section \ref{sec:ml:train}.
Even a perfect probabilistic model, however, could well exhibit this  behaviour due to a counter-intuitive property of many high-dimensional random variables that causes the mode of the distribution to be very different from \emph{typical} samples, which have a log-probability close to the entropy of the distribution.
(See \citet{Cover2006} for a detailed discussion of \emph{typicality} from an information theory perspective).
\citet{eikema-aziz-2020-map} recognise this issue in the context of NMT and tackle it by applying \emph{Minimum Bayes Risk} (MBR) inference \citep{Goel2000}.

Minimum Bayes Risk seeks to generate a translation which is maximally similar, according to a metric such as BLEU or METEOR \citep{denkowski-lavie-2011-meteor}, to other translations sampled from the model itself, each weighted according to its probability.
The intuition is that the generated translation will belong to a high-probability cluster of similar candidate translations; highly abnormal translations such as the empty sentence will be excluded.
\citet{Eikema2019} report improvements over beam search on the low-resource language pairs of the FLoRes dataset (Nepali--English and Sinhala--English) \citep{guzman-etal-2019-flores} while they lose some accuracy on English--German.
They also evaluate inference though ancestral sampling, the simplest and theoretically least biased inference technique, but they found that it performs worse than both beam search and MBR.

\emph{Energy-based models} (EBMs) \citep{lecun2006tutorial} are alternative representations of a probability distribution which can be used for inference.
An ERM of a random variable (a whole sentence, in our case) 
is a scalar-valued \emph{energy function}, 
implemented as a neural network, which represents an unnormalised log-probability.
This lack of normalisation means that only probability ratios between two sentences can be computed efficiently, for this reason training and sampling from EBMs requires a proposal distribution to generate reasonably good initial samples to be re-ranked by the model, in the context of MT this proposal distribution is a conventional autoregressive NMT model.
\citet{naskar2020energybased} define source-conditional 
or joint 
EBMs trained on ancestral samples from an autoregressive NMT model using a reference-based metric (e.g.~BLEU).
During inference they apply the EBM to re-rank a list of $N$ ancestral samples from the autoregressive NMT model.
This approximates MAP inference on a probability distribution that tracks the reference-based metric, which would not give high weight to abnormal translations such as the empty sentence.
They report improvements on multiple language pairs, in particular for medium-resource and low-resource language pairs such as Romanian,  Nepali, and Sinhala to English.

Reranking has been also applied under the generalised noisy channel model initially developed for SMT \citep{koehn-etal-2003-statistical}, where translations are scored not just under the probability of the target conditional on the source (direct model) but also under the probability of the source conditional on the target (channel model) and the unconditional probability of the target (language model prior), combined by a weighted sum of their logarithms.
This reranking can be applied at sentence level on a set of candidate translations generated by the direct model by conventional beam search \citep{chen-etal-2020-facebook} or at token level interleaved with beam search \citep{bhosale-etal-2020-language}, resulting in improvements in multiple language pairs including low-resource ones.

\subsection{Rule-based approaches}
\label{sec:ml:rulebased}

Rule-based machine translation (RBMT) consists in analysing and transforming a source text into a translation by applying a set of hand-coded linguistically motivated rules.
This was the oldest and the most common paradigm for machine translation before being largely supplanted by corpus-based approaches such as phrase-based statistical machine translation and neural machine translation, which usually outperform it on both accuracy and fluency, especially when translating between language pairs where at least one language is high-resource, such as English.
However, rule-based techniques can still be successfully applied to the task of translation between closely related languages.

Modern implementations, such as the \textit{Apertium} system \citep{forcada2011apertium, forcada-tyers-2016-apertium, khanna2021recent}, use lexical translation and \textit{shallow transfer} rules that avoid full parsing and instead exploit the similarities between the source and target language to restructure a sentence into its translation.
This approach has been applied to various language pairs, especially in the Western Romance and the South Slavic sub-families.
NMT approaches tend to have better fluency than RBMT but they can produce hallucinations in low-resource settings where RBMT can instead benefit from lexical translation with explicit bilingual dictionaries, thus a line of research has developed that attempts to combine both approaches.
For instance \citet{sanchez-cartagena-etal-2020-multi} used multi-source Transformer and deep-GRU \citep{miceli-barone-etal-2017-deep} models to post-edit translations produced by a RBMT system for the Breton–French language pair.

One of the main drawbacks of RBMT is that it requires substantial language-specific resources and expertise which might not be available for all low-resource languages.
See Section~\ref{sec:linguistic} for a discussion of other methods to use various linguistic resources that might be more readily available.
\section{Evaluation}
\label{sec:eval}

As researchers build different MT systems in order to try out new ideas, how do they know whether one is better than another? If a system developer wants to deploy an MT system, how do they know which is the best?
Answering these questions is the  goal of MT \emph{evaluation} -- to provide a quantitative estimate of the quality of an MT system's output. MT evaluation is a difficult problem, since there can be many possible correct translations of a given
source sentence. The intended use is an important consideration in evaluation; if translation is mainly for
assimilation (gisting) then adequacy is of primary importance and errors in fluency can be tolerated; but if the
translation is for dissemination (with post-editing) then errors in meaning can be corrected, but the translation should be as close to a publishable form as possible.
Evaluation is not specific to low-resource MT; it is a problem for all types of MT research, but low-resource language pairs can present specific difficulties for evaluation.

Evaluation can either be \emph{manual} (using human judgements) or \emph{automatic} (using software). Human judgements are generally considered the ``gold standard'' for MT evaluation because, ultimately, the translation is intended to be consumed by humans. The annual WMT shared tasks have employed human evaluation every year since they started in 2006, and the organisers argue that \citep{callison-burch-etal-2007-meta}:
\begin{quote}
While
automatic measures are an invaluable tool for the
day-to-day development of machine translation systems, they are an imperfect substitute for human
assessment of translation quality. 
\end{quote}
Human evaluation is of course much more time-consuming (and therefore expensive) than automatic evaluation, and so can 
only be used to compare a small number of variants; with most of the system selection performed by automatic evaluation. For low-resource MT, the potential difficulty with human evaluation is connecting the researchers with the evaluators. Some low-resource languages have very small language communities, so the pool of potential evaluators is small, whilst in other cases the researchers may not be well connected with the language community -- in that case the answer should be for the researchers
to engage more with the community \citep{nekoto-etal-2020-participatory}.

Most of the evaluation in MT is performed using automatic metrics, but when these metrics are developed they need to be
validated against human judgements as the gold standard. An important source of gold standard data for validation of 
metrics is the series of WMT metrics shared tasks \citep{freitag-EtAl:2021:WMT}. However this data covers the language pairs
used in the WMT news tasks, whose coverage of low-resource languages is limited to those listed in Table~\ref{tab:shared}. It 
is unclear how well conclusions about the utility of metrics will transfer from high-resource languages to low-resource languages.

Automatic metrics are nearly always reference-based, in other words they work by comparing the MT hypothesis with 
a human-produced reference. This means that references need to be available for the chosen language pair, they should
be of good quality, and ideally should be established benchmarks used by the research community. Such references
are in short supply for low-resource language pairs (Section~\ref{sec:data}), and when available may be in the wrong domain,
small, or of poor quality.

Looking at the automatic metrics in common use in current MT research, we see two main types of metrics being used in current research:  
string-based metrics (e.g.~BLEU \citep{papineni-etal-2002-bleu, post-2018-call}, ChrF \citep{popovic_chrf_2015} etc.) 
and embedding-based  metrics (e.g.~BERTscore \cite{zhang_bertscore_2019}, COMET \citep{rei-etal-2020-comet}, BLEURT \cite{sellam-etal-2020-bleurt} etc.). The string-based metrics are ``low-resource metrics'', since they do not require any resources beyond tokenisation/segmentation, whereas the embedding-based metrics require more significant resources such as pre-trained sentence embeddings, and often need human
judgements for fine-tuning. The embedding-based metrics could be considered successors to metrics like 
METEOR \cite{denkowski-lavie-2014-meteor}, which uses synonym list to improve matching between hypothesis and reference.

Recent 
comparisons \cite{freitag-EtAl:2021:WMT,kocmi-EtAl:2021:WMT} have suggested that embedding-based metrics have superior performance, 
in other words that they correlate better with human judgements than string-based metrics. However, since embedding-based metrics generally rely on sentence
embeddings, they only work when such embeddings are available, and if they are fine-tuned on human evaluation data, they may not
perform as well when such data is not available. For instance COMET is based on the XML-R embeddings \citep{conneau-etal-2020-unsupervised}, so can support the 100 languages supported by XML-R, but will not give reliable results for unsupported languages.

String-based metrics (such as BLEU and ChrF) will generally support any language for reference-based automatic 
evaluation. BLEU has been in use in MT evaluation for many years, and its benefits and limitations are well-studied; ChrF has a much shorter history, but recent comparisons suggest that it performs better than BLEU (see references above) and its use of character $n$-grams
probably make it more suited to the morphological complexity found in many low-resource languages. A recent debate in 
the MT evaluation literature has been about the ability for automatic metrics to discern differences in high-quality
MT systems \citep[Figure 6]{ma-etal-2019-results}. However in low-resource MT, we may be faced with the opposite problem, i.e.\ metrics
may be less reliable when faced with several low-quality systems \cite{fomicheva-specia-2019-taking}.

In conclusion, for evaluation of low-resource MT, we recommend human evaluation as the gold standard, but where automatic evaluation is used, to be especially wary of the lack of calibration of metrics and the potential unreliability of test 
sets for low-resource language pairs. 
\section{Shared Tasks}
\label{sec:cases}


MT is a big field and many interesting papers are published all the time. Because of the variety of language pairs, toolkits and settings, it can be difficult to determine what research will have an impact beyond the published experiments. Shared tasks provide an opportunity to reproduce and combine research, while keeping the training and testing data constant.

System description papers can offer valuable insights into how research ideas can transfer to real gains when aiming to produce the best possible system with the data available. Whilst standard research papers often focus on showing that the technique or model proposed in the paper is effective, the incentives for system descriptions are different; authors are concerned with selecting the techniques (or most commonly the combination of techniques) that work best for the task. System descriptions therefore contain a good reflection of the techniques that researchers believe will work (together with their comparison) in standardised conditions.
The difficulty with system descriptions is that the submitted systems are often potpourris of many different techniques, organised in pipelines with multiple steps, a situation that rarely occurs in research papers presenting individual approaches. In light of this, it is not always easy to pinpoint exactly which techniques lead to strongly performing systems, and it is often the case that similar techniques are used by both leading systems and those that perform less well. Moreover, the papers do not tend to provide an exhaustive and systematic comparison of different techniques due to differences in implementation, data processing and hyper-parameters. We also note that the evaluation of shared tasks normally focuses on quality alone, although a multi-dimensional analysis may be more appropriate
\citep{ethayarajh_utility_2020}, and that even if the task has manual evaluation, there is still debate about the 
best way to do this \citep{freitag_experts_2021}.

Apart from the system descriptions, an important output of shared tasks is the publication of standard training sets and test sets (Section~\ref{sec:data}). These can be used in later research, and help to raise the profile of the language pair for MT research.

In this section we survey the shared tasks that have included low-resource language pairs, and we draw common themes from the corresponding sets of system description papers, putting into perspective the methods previously described in this survey.
Rather than attempting to quantify the use of different techniques \textit{à la} \citet{libovicky2021blog0724}\footnote{\url{https://jlibovicky.github.io/2021/07/24/MT-Weekly-The-Wisdom-of-the-WMT-Crowd}}, we aim to describe how the most commonly used techniques are exploited, particularly for high-performing systems, 
providing some practical advice to training systems for low-resource language pairs. We begin with a brief description of shared tasks featuring low-resource pairs (Section~\ref{sec:lr-shared}), before surveying techniques commonly used (Section~\ref{sec:lr-case-study}).



\subsection{Low-resource MT in Shared Tasks}
\label{sec:lr-shared}
There are many shared tasks that focus on MT, going all the way back to the earliest WMT shared
task \citep{koehn_manual_2006}. However they have tended to focus on well-resourced European languages and Chinese. Tasks specifically for low-resource MT are fairly new, coinciding with the recent interest in expanding MT to a
larger range of language pairs.

We choose to focus particularly on shared tasks run by WMT (WMT Conference on Machine Translation), IWSLT (International Conference on Spoken Language Translation),
WAT (Workshop on Asian Translation)  and LowResMT (Low Resource Machine Translation). In Table~\ref{tab:shared}, we list the shared MT task that have focused on low-resource pairs. In addition to the translation tasks, we should mention that the corpus filtering task at WMT has specifically addressed low-resource MT \citep{koehn-etal-2019-findings,koehn-EtAl:2020:WMT}.

\begin{table}[ht!]
    \begin{center}
    \small
    \begin{tabular}{cp{5.5cm}p{6cm}}
    \toprule
    Year & Task name and reference & Language pairs \\
    \midrule
    2018 & IWSLT \citep{iwslt2018overview} & Basque--English\\
    2018 & WAT Mixed domain  \citep{nakazawa-etal-2018-overview} & Myanmar--English \\
    2019 & WAT Mixed domain  \citep{nakazawa-etal-2019-overview} & Myanmar--English and Khmer--English\\
    2019 & WAT Indic  \citep{nakazawa-etal-2019-overview} & Tamil--English\\
    2019 & WMT news \citep{barrault-etal-2019-findings} & Kazakh--English and Gujarati--English  \\
    2019 & LowResMT \citep{ojha-etal-2020-findings} & \{Bhojpuri, Latvian, Magahi and Sindhi\} --English \\
    2020 & WMT news \citep{barrault-EtAl:2020:WMT1} & \{Tamil, Inuktitut,
    Pashto and Khmer\} --English \\
    
    2020 & WMT Unsupervised and very low resource \cite{fraser-2020-findings} & Upper Sorbian--German \\ 
    2020 & WMT Similar language  \cite{barrault-EtAl:2020:WMT1} & Hindi--Marathi \\
    2020 & WAT Mixed domain \citep{nakazawa-etal-2020-overview} &  Myanmar--English and Khmer--English\\
    2020 & WAT Indic   \citep{nakazawa-etal-2020-overview} & Odia--English \\
    2021 & AmericasNLP \cite{mager-etal-2021-findings} &   Ten indigenous languages of Latin America, to/from Spanish \\
    2021 & WAT News Comm \citep{nakazawa-etal-2021-overview} & Japanese--Russian \\
    2021 & WAT Indic \citep{nakazawa-etal-2021-overview} & Ten Indian languages, to/from English \\
    2021 & LowResMT \citep{mtsummit-2021-technologies} & Taiwanese Sign Language--Trad. Chinese, Irish--English and Marathi--English \\
    2021 & WMT News \citep{findings2021wmt} & Hausa--English and Bengali--Hindi \\
    2021 & WMT Unsupervised and very low resource \citep{libovick-fraser:2021:WMT1} & Chuvash--Russian and Upper Sorbian--German \\
    2021 & WMT  Large-Scale Multilingual
MT \citep{wenzek2021wmt} & FLORES-101 2 small and 1 large task (10k pairs) \\
    \bottomrule
    \end{tabular}
    \caption{Shared tasks that have included low-resource language pairs} 
    \label{tab:shared}
 
    \end{center}
\end{table}

\subsection{Commonly used Techniques}\label{sec:lr-case-study}
In this section, we review the choices made by participants to shared tasks for low-resource MT, focusing on those techniques that are particularly widespread, those that work particularly well and the choices that are specific to particular languages or language families. We describe these choices in an approximately step-by-step fashion: starting with data preparation (Section~\ref{sec:case-study:data-prep}) and data processing (Section~\ref{sec:case-study:data-proc}), then proceeding to model architecture choices (Section~\ref{sec:case-study:model}), exploiting additional data, including backtranslation, pretraining and multilinguality (Section~\ref{sec:case-study:additional-data}) and finally looking at model transformation and finalisation, including ensembling, knowledge distillation and fine-tuning (Section~\ref{sec:case-study:finalisation}).

\subsubsection{Data preparation}\label{sec:case-study:data-prep}

An important initial step to training an NMT model is to identify available data (See Section~\ref{sec:data}) and to potentially filter it depending on the noisiness of the dataset and how out-of-domain it is or to use an alternative strategy to indicate domain or data quality (i.e.~tagging). So what choices do participants tend to make in terms of using (or excluding) data sources, filtering and cleaning of data and using meta-information such as domain tags?

\paragraph{Choice of data} We focus on constrained submissions only (i.e.~where participants can only use the data provided by the organisers), so most participants use all available data. Hesitancy can sometimes be seen with regards to web-crawled data (other than WMT newscrawl, which is generally more homogeneous and therefore of better quality), some choosing to omit the data \citep{singh-2020-adobe} and others to filter it for quality \citep{chen-etal-2020-facebook,li-EtAl:2020:WMT}. It is very unusual to see teams do their own crawling (\citet{hernandez-nguyen:2020:WMT} is a counter-example); teams doing so run the risk of crawling data that overlaps with the development set or one side of the test set.
\citet{tran2021wmt} successfully mined an extra million sentences pairs of Hausa-English data from the allowed monolingual data, helping them win the 2021 task. 

\paragraph{Data cleaning and filtering} Although not exhaustively reported, many of the submissions apply some degree of data cleaning and filtering to the parallel and monolingual data. In its simplest form, this means excluding sentences based on their length (if too long) and the ratio between the lengths of parallel sentences (if too different). Some teams also remove duplicates (e.g.~\citet{li-EtAl:2019:WMT2}). More rigorous cleaning includes eliminating sentences containing fewer than a specified percentage of alpha-numeric characters in sentences (depending on the language's script), those identified as belonging to another language (e.g.~using language identification) or those less likely to belong to the same distribution as the training data (e.g.~using filtering techniques such as Moore-Lewis \citep{moore-lewis-2010-intelligent}). Data filtering is also a commonly used technique for backtranslation data (see the paragraph on data augmentation below), often using similar filtering techniques such as dual conditional cross-entropy filtering \citep{junczys-dowmunt-2018-dual} to retain only the cleanest and most relevant synthetic parallel sentences. Unfortunately the effect of data filtering is rarely evaluated, probably because it would involve expensive re-training.

\paragraph{Data tagging}
Some teams choose to include meta-information in their models through the addition of pseudo-tokens. For example, \citet{dutta-EtAl:2020:WMT} choose to tag sentences according to their quality for the Upper Sorbian--German task, this information being provided by the organisers. Domain tagging (i.e.~indicating the type of data), which can be useful to indicate whether data is in-domain or out-of-domain was used by \citet{chen-etal-2020-facebook}, one of the top-scoring systems for Tamil--English. For the Basque--English task, \citet{scherrer-helsinki-2018-IWSLT} find that using domain tags gives systematic improvements over not using them, and \citet{knowles-EtAl:2020:WMT2} come to the same conclusion when translating into Inuktitut.


\subsubsection{Data pre-processing}\label{sec:case-study:data-proc}

There is some variation in which data pre-processing steps are used. For example, it has been shown that for high-resource language pairs such as Czech--English, it is not always necessary to applying tokenisation and truecasing steps \citep{bawden-etal-2019-university} before apply subword segmentation. We do not observe a clear pattern, with many systems applying all steps, and some excluding tokenisation (\citealt{wu-EtAl:2020:WMT1} for Tamil) and truecasing. Among the different possible pre-processing steps, we review participants choices concerning tokenisation, subword segmentation and transliteration/alphabet mapping (relevant when translating between languages that use different scripts).

\paragraph{Tokenisation} If a tokeniser is used before subword segmentation, it is common for it to be language-specific, particularly for the low-resource language in question. For example IndicNLP\footnote{\href{https://github.com/anoopkunchukuttan/indic\_nlp\_library}{https://github.com/anoopkunchukuttan/indic\_nlp\_library}} \citep{kunchukuttan2020indicnlp} is widely used for Indian languages (e.g.~for the shared tasks involving Gujarati and Tamil), and many of the Khmer--English submissions also used Khmer-specific tokenisers. For European languages, the Moses tokeniser \citep{koehn-etal-2007-moses} remains the most commonly used option. 

\paragraph{Subword segmentation} All participants perform some sort of subword segmentation, with most participants using either sentencepiece \citep{kudo-richardson-2018-sentencepiece}\footnote{\href{https://github.com/google/sentencepiece}{https://github.com/google/sentencepiece}} or subword\_nmt toolkits \citep{sennrich2016neural}.\footnote{\href{https://github.com/rsennrich/subword-nmt}{https://github.com/rsennrich/subword-nmt}} Even though the BPE toolkit is not compatible with the Abugida scripts used for Gujarati, Tamil
and Khmer (in these scripts, two unicode codepoints can be used to represent one glyph), we only found one
group who modified BPE to take this into account \citep{shi-etal-2020-oppos}. BPE-dropout \citep{Provilkov2020-zw}, a regularisation method,  was found to be useful by a number of teams \citep{knowles-EtAl:2020:WMT1,libovick-EtAl:2020:WMT,chronopoulou-EtAl:2020:WMT}.

The size of the subword vocabulary is often a tuned parameter, although the range of different values tested is not always reported. Surprisingly, there is significant variation in the subword vocabulary sizes used, and there is not always a clear pattern. Despite the low-resource settings, many of the systems use quite large subword vocabularies (30k-60k merge operations). There are exceptions: a large number of the systems for Tamil--English use small vocabularies (6k-30k merge operations), which may be attributed to the morphologically rich nature of Tamil coupled with the scarcity of data.

Joint subword segmentation is fairly common. Its use is particularly well motivated when the source and target languages are similar where we may expect to see a high amount of lexical overlap (e.g.~for the similar language shared tasks such as Upper Sorbian--German) and when `helper languages' are used to compensate for the low-resource scenario (e.g.~addition of Czech and English data). However, it is also used in some cases even where there is little lexical overlap, for example for Tamil--English, where the languages do not share the same script, including by some of the top-scoring systems \citep{shi-etal-2020-oppos,wu-etal-2018-beyond}. Although few systematic studies are reported, one hypothesis could be that even if different scripts are used there is no disadvantage to sharing segmentation; it could help with named entities and therefore reducing the overall vocabulary size of the model \citep{ding-etal-2019-call}.

A few works explore alternative morphology-driven segmentation schemes, but without seeing any clear advantage: \citet{scherrer-grnroos-virpioja:2020:WMT} find that, for Upper-Sorbian--German, Morfessor can equal the performance of BPE when tuned correctly (but without surpassing it), whereas \citet{sanchez-prompsit-2018-IWSLT} find gains for Morfessor over BPE. \citet{dhar-etal-2020-linguistically} have mixed results for Tamil--English when comparing linguistically motivated subword units compared to the use of statistics-based sentencepiece \citep{kudo-richardson-2018-sentencepiece}.

\paragraph{Transliteration and alphabet mapping}

Transliteration and alphabet mapping has been principally used in the context of exploiting data from related languages that are written in different scripts. 
This was particularly present for translation involving Indian languages, which often have their own script. For the Gujarati--English task, many of the top systems used Hindi--English data (see below the paragraph on using other language data) and performed alphabet mapping into the Gujarati script \citep{li-etal-2019-niutrans,bawden-etal-2019-university,dabre-EtAl:2019:WMT}.
For Tamil--English, \citet{goyal-EtAl:2020:WMT} found that when using Hindi in a multilingual setup, it helped for Hindi to be mapped into the Tamil script for the Tamil$\rightarrow$English direction, but did not bring improvements for English$\rightarrow$Tamil.
Transliteration was also used in the Kazakh--English task, particularly with the addition of Turkish as higher-resourced language. \citet{toral-EtAl:2019:WMT}, a top-scoring system, chose to cyrillise Turkish to increase overlap with Kazakh, whereas \citet{briakou-carpuat:2019:WMT} chose to romanise Kazakh to increase the overlap with Turkish, but only for the Kazakh$\rightarrow$English direction.

\subsubsection{Model architectures and training}\label{sec:case-study:model}

The community has largely converged on a common architecture, the Transformer \citep{vaswani2017attention}, although differences can be observed in terms of the number of parameters in the model and certain training parameters. It is particularly tricky to make generalisations about model and training parameters given the dependency on other techniques used (which can affect how much data is available). However a few generalisations can be seen, which we review here.

\paragraph{SMT versus NMT}
There is little doubt that NMT has overtaken SMT, even in the low-resource tasks. The majority of submissions use neural MT models and more specifically transformers (rather than recurrent models). Some teams compare SMT and NMT  \citep{dutta-EtAl:2020:WMT,sen-EtAl:2019:WMT} with the conclusion that NMT is better when sufficient data is available, including synthetic data. Some teams use SMT only for back-translation, on the basis that SMT can work better (or at least be less susceptible to hallucinating) on the initial training using a very small amount of parallel data. For SMT systems, the most commonly used toolkit is Moses \citep{koehn-etal-2007-moses}, whereas there is a little more variation for NMT toolkits; the most commonly used being Fairseq \citep{ott-etal-2019-fairseq}, Marian \citep{mariannmt2018}, OpenNMT \citep{opennmt2017} and Sockeye \citep{hieber-etal-2020-sockeye}.

\paragraph{Model size}

Although systematic comparisons are not always given, some participants did indicate that architecture size was a tuned parameter \citep{chen-etal-2020-facebook}, although this can be computationally expensive and therefore not a possibility for all teams. The model sizes chosen for submissions varies, and there is not a clear and direct link between size and model performance. However there are some general patterns worth commenting on.
While many of the baseline models are small (corresponding to transformer-base or models with fewer layers), a number of high-scoring teams found that it was possible to train larger models (e.g.~deeper or wider) as long as additional techniques were used, such as monolingual pretraining \citep{wu-EtAl:2020:WMT1} or additional data from other languages in a multilingual setup or after synthetic data creation through pivoting \citep{li-etal-2019-niutrans} through a higher-resource language or backtranslation \citep{chen-etal-2020-facebook,li-etal-2019-niutrans}. For example the Facebook AI team \citep{chen-etal-2020-facebook}, who fine-tuned for model architecture, started with a smaller transformer (3 layers and 8 attention heads) for their supervised English$\rightarrow$Tamil baseline, but were able to increase this once backtranslated data was introduced (to 10 layers and 16 attention heads). Although some systems perform well with a transformer-base model (\citealt{bawden-etal-2019-university} for Tamil--English), many of the best systems use larger models, such as the transformer-big \citep{hernandez-nguyen:2020:WMT,kocmi:2020:WMT,bei-etal-2019-gtcom,wei-EtAl:2020:WMT1,chen-etal-2020-facebook}.

\paragraph{Alternative neural architectures}
Other than variations on the basic transformer model, there are few alternative architectures tested. 
\citet{wu-EtAl:2020:WMT1} tested the addition of dynamic convolutions to the transformer model following \citet{wu-pay-less-2019}, which they used along with other wider and deep transformers in model ensembles. However they did not compare the different models.
Another alternative form of modelling tested by several teams was factored representations (see Section~\ref{sec:linguistic_factored}).  \citet{dutta-EtAl:2020:WMT} explored the addition of lemmas and part-of-speech tags for Upper-Sorbian--German but without seeing gains, since the morphological tool used was not adapted to Upper-Sorbian. For Basque--English, \citet{SRPOL-UEDIN-iwslt-2018} find that source factors indicating the language of the subword can help to improve the baseline system.

\paragraph{Training parameters}
Exact training parameters are often not provided, making comparison difficult. Many of the participants do not seem to choose training parameters that are markedly different from the higher-resource settings \citep{zhang-EtAl:2020:WMT,wu-EtAl:2020:WMT1}. 


\subsubsection{Using additional data}\label{sec:case-study:additional-data}
Much of this survey has been dedicated to approaches for the exploitation of additional resources to compensate for the lack of data for low-resource language pairs: monolingual data (Section~\ref{sec:monolingual}), multilingual data (Section~\ref{sec:multilingual}) or other linguistic resources (Section~\ref{sec:linguistic}). In shared tasks, the following approaches have been shown to be highly effective to boosting performance in low-resource scenarios.

\paragraph{Backtranslation}
The majority of high-performing systems carry out some sort of data augmentation, the most common being backtranslation, often used iteratively, although forward translation is also used \citep{shi-etal-2020-oppos,chen-etal-2020-facebook,zhang-EtAl:2020:WMT,wei-EtAl:2020:WMT1}. For particularly challenging language pairs (e.g.~for very low-resource between languages that are not very close), it is important for the initial model that is used to produce the backtranslations to be of sufficiently high quality. For example, some of the top Gujarati--English systems employed pretraining before backtranslation to boost the quality of the initial model \citep{bawden-etal-2019-university,bei-etal-2019-gtcom}. 
Participants do not always report the number of iterations of backtranslations performed, however those that do often cite the fact that few improvements are seen beyond two iterations
\cite{chen-etal-2020-facebook}.
Tagged backtranslation, whereby a pseudo-token is added to sentences that are backtranslated to distinguish then from genuine parallel data have previously shown to provide improvements \citep{caswell-etal-2019-tagged}. Several participants report gains thanks to the addition of backtranslation tags \cite{wu-EtAl:2020:WMT1,chen-etal-2020-facebook,knowles-EtAl:2020:WMT2}, although \citet{goyal-EtAl:2020:WMT} find that tagged backtranslation under-performs normal backtranslation in a multilingual setup for Tamil--English.



\paragraph{Synthetic data from other languages}
A number of top-performing systems successfully exploit parallel corpora from related languages.
The two top-performing systems for Gujarati--English use a Hindi--English parallel corpus to create synthetic Gujarati--English data \citep{li-etal-2019-niutrans,bawden-etal-2019-university}. Both exploit the fact that there is a high degree of lexical overlap between Hindi and Gujarati once Hindi has been transliterated into Gujarati script. \citet{li-etal-2019-niutrans} choose to transliterate the Hindi side and then to select the best sentences using cross-entropy filtering, whereas \citet{bawden-etal-2019-university} choose to train a Hindi$\rightarrow$Gujarati model, which they use to translate the Hindi side of the corpus.
Pivoting through a higher-resource related language was also found to be useful for other language pairs: for Kazakh--English, Russian was the language of choice \citep{li-etal-2019-niutrans,toral-EtAl:2019:WMT,dabre-EtAl:2019:WMT,budiwati-EtAl:2019:WMT}, for Basque--English, Spanish was used as a pivot \citep{scherrer-helsinki-2018-IWSLT,sanchez-prompsit-2018-IWSLT}, which was found to be more effective than backtranslation by \citet{scherrer-helsinki-2018-IWSLT}, and was found to benefit from additional filtering by \citet{sanchez-prompsit-2018-IWSLT}.

\paragraph{Transfer-learning using language modelling objectives}
The top choices of language modelling objectives are \textit{mBART} \citep{liu2020multilingual} (used by \citet{chen-etal-2020-facebook} and \citet{bawden-EtAl:2020:WMT1} for Tamil--English), \textit{XLM} \citep{lample2019cross} (used by \citet{bawden-etal-2019-university} for Gujarati--English, by \citet{laskar-EtAl:2020:WMT} for Hindi--Marathi, and by \citet{kvapilkov-kocmi-bojar:2020:WMT} and \citet{dutta-EtAl:2020:WMT} for Upper-Sorbian--German), and \textit{MASS} \citep{song2019mass} (used by \citet{li-EtAl:2020:WMT} and \citet{singh-singh-bandyopadhyay:2020:WMT} for Upper Sorbian--German). Some of the top systems used these language modelling objectives, but their use was not across the board, and pretraining using translation objectives was arguably more common. Given the success of pretrained models in NLP, this could be surprising. A possible explanation for these techniques not being used systematically is that they can be computationally expensive to train from scratch and the constrained nature of the shared tasks means that the participants are discouraged from using pretrained language models.


\paragraph{Transfer learning from other MT systems}
Another commonly used technique used by participants was transfer learning 
 involving other language pairs. 
Many of the teams exploited a high-resource related language pair. For example, for Kazakh--English, pretraining was done using Turkish--English \citep{briakou-carpuat:2019:WMT} and Russian--English \citep{kocmi-bojar:2019:WMT}, \citet{dabre-EtAl:2019:WMT} pretrained for Gujarati--English using Hindi--English, and Czech--German was used to pretrain for Upper-Sorbian--German \citep{knowles-EtAl:2020:WMT1}.

An alternative but successful approach was to use a high-resource but not necessarily related language pair. For example, the CUNI systems use Czech--English to pretrain Inuktitut \citep{kocmi:2020:WMT} and Gujarati \citep{kocmi-bojar:2019:WMT}, and \citet{bawden-EtAl:2020:WMT1} found pretraining on English--German to be as effective as mBART training for Tamil--English.
Finally, a number of teams opted for multilingual pretraining, involving the language pair in question and a higher-resource language or several higher-resource languages. \citet{wu-EtAl:2020:WMT1} use the mRASP approach: a universal multilingual model involving language data for English to and from Pashto, Khmer, Tamil, Inuktitut, German and Polish, which is then fine-tuned to the individual low-resource language pairs.

\paragraph{Multilingual models}
Other than the pretraining strategies mentioned just above, multilingual models feature heavily in shared task submissions. The overwhelmingly most common framework used was the universal encoder-decoder models as proposed by \citet{johnson2017google}. Some participants chose to include select (related) languages. \citet{SRPOL-UEDIN-iwslt-2018} use Spanish to boost Basque--English translation and find that the addition of French data degrades results. \citet{goyal-sharma:2019:WMT} add Hindi as an additional encoder language for Gujarati--English and for Tamil--English, they test adding Hindi to either the source or target side depending on whether Tamil is the source or target language \citep{goyal-EtAl:2020:WMT}. Other participants choose to use a larger number of languages. \citet{zhang-EtAl:2020:WMT} train a multilingual system on six Indian languages for Tamil--English and \citet{hokamp-glover-gholipourghalandari:2019:WMT} choose to train a multilingual model on all WMT languages for Gujarati--English (coming middle in the results table).
Upsampling the lower-resourced languages in the multilingual systems is an important factor, whether the multilingual system is used as the main model or for pretraining \citep{zhang-EtAl:2020:WMT,wu-EtAl:2020:WMT1}. Recent approaches has seen success using more diverse and even larger numbers of languages. 
\citet{tran2021wmt} train a model for 14 diverse language directions, winning 10 of them (although their system 
is unconstrained). Their two models, many to English and English to many, used a Sparsely Gated Mixture-of-Expert (MoE) models \citep{lepikhin2020gshard}. The MoE strike a balance between allowing high-resource directions to benefit from increased model capacity, while also allowing transfer to
low-resource directions via shared capacity.
Microsoft's winning submission to the 2021 large scale multilingual task~\citep{jian2021wmt} covered 10k language pairs across the FLORES-101 data set. They use the public available DeltaLM-Large model \citep{ma2021deltalm}, a multilingual pre-trained encoder-decoder model, and apply progressive learning~\citep{zhang2020learning} (starting training with 24 encoder layers and adding 12 more) and iterative back-translation.

%


\subsubsection{Model transformation and finalisation}\label{sec:case-study:finalisation}

Additional techniques, not specific to low-resource MT, are often applied in the final stages of model construction, and they can provide significant gains to an already trained model. We regroup here knowledge distillation (which we consider as a sort of model transformation) and both model combination and fine-tuning (which can be considered model finalisation techniques).

\paragraph{Knowledge distillation}
Knowledge distillation is also a frequently used technique and seems to give minor gains, although is not as frequently used as backtranslation or ensembling. Knowledge distillation~\citep{kim-rush-2016-sequence} leverages a large teacher model to train a student model. The teacher model is used to translate the training data, resulting in synthetic data on the target side. 
A number of teams apply this iteratively, in combination with backtranslation \citep{xia-EtAl:2019:WMT} or fine-tuning \citep{li-etal-2019-niutrans}. \citet{bei-etal-2019-gtcom} mix knowledge distillation data with genuine and synthetic parallel data to train a new model to achieve gains in BLEU.

\paragraph{Model combination}
Ensembling is the combination of several independently trained models and is used by a large number of participants to get gains over single systems. Several teams seek to create ensembles of diverse models, including deep and wide ones. For example \citet{wu-EtAl:2020:WMT1} experiment with ensembling for larger models (larger feed forward dimension and then deeper models), including using different sampling strategies to increase the number of different models.
Ensembling generally leads to better results but not always. \citet{wu-EtAl:2020:WMT1}  found that a 9-model ensemble was best for Khmer and Pashto into English, but they found that for English into Khmer and Pashto, a single model was best.
A second way of combining several models is to use an additional model to rerank $n$-best hypothesis of an initial model. \citet{libovick-EtAl:2020:WMT} attempted right-to-left rescoring (against the normally produced left-to-right hypothesis), but without seeing any gains for Upper-Sorbian--German. \citet{chen-etal-2020-facebook} test noisy channel reranking for Tamil--English, but without seeing gains either, although some gains are seen for Inuktitut$\rightarrow$English, presumably because of the high-quality monolingual news data available to train a good English language model.

\paragraph{Fine-tuning}
Mentioned previously in the context of pretraining, fine-tuning was used in several contexts by a large number of teams. It is inevitably used after pretraining on language model objectives or on other language pairs (see above) to adapt the model to the language direction in question. It is also frequently used on models trained on backtranslated data, by fine-tuning on genuine parallel data~\citep{sanchez-cartagena-etal-2019-universitat}. A final boost used by a number of top-systems is achieved through fine-tuning to the development set \citep{shi-etal-2020-oppos,chen-etal-2020-facebook,zhang-EtAl:2020:WMT,wei-EtAl:2020:WMT1}. This was choice not made by all teams, some of which chose to keep it as a held-out set, notably to avoid the risk of overfitting.




\section{Conclusion}
In the previous section (Section~\ref{sec:cases}), we saw that even if shared tasks rarely offer definitive answers, they do give a good indication of what combination of techniques can deliver state-of-the-art systems for particular language pairs. This look at what methods are commonly used in practice hopefully provides some perspective on the research we have covered in this survey.

In this survey we have  looked at the entire spectrum of scientific effort in the field: from data sourcing and creation (Section~\ref{sec:data}), leveraging all types of available data, monolingual data (Section~\ref{sec:monolingual}), multilingual data (Section~\ref{sec:multilingual}) and other linguistic resources (Section~\ref{sec:linguistic}), to improving model robustness, training and inference (Section~\ref{sec:ml}), and finally evaluating the results (Section~\ref{sec:eval}).

Thanks to large-scale and also more focused efforts to identify, scrape and filter parallel data from the web, some language pairs have moved quickly from being considered low-resource to now being considered more medium-resourced (e.g.~Romanian--English, Turkish--English and Hindi--English). The ability of deep learning models to learn from monolingual data~\citep{sennrich2016improving,cheng2016semi,NIPS2016_5b69b9cb} and related languages  \citep{liu2020multilingual, lample2019cross, pan2021contrastive} has had a big impact on this field.
However state-of-the-art systems for high-resourced language pairs like German-English and Chinese-English still seem to 
require huge amounts of data \cite{akhbardeh-EtAl:2021:WMT}, far more than a human learner.

So how far have we come, and can we quantify the difference between the performance on high- and low- resource
language pairs? Examining recent comparisons reveals a mixed picture, and shows the difficulty of trying to compare results across language pairs. In the WMT21 news task, we can compare the German-English (nearly 100M sentences of parallel training data) with the Hausa-English (just over 0.5M sentences parallel) tasks. The direct assessment (DA) scores of the best systems are similar: German$\rightarrow$ English (71.9),
Hausa$\rightarrow$ English (74.4), English$\rightarrow$German (83.3) and English$\rightarrow$Hausa (82.7), on a 0--100 scale. In fact, for both the out-of-English pairs, the evaluation did not find the best system to be statistically significantly worse than the human translation. In the WMT21 Unsupervised and Low-resource task 
\citep{libovick-fraser:2021:WMT1}, BLEU scores of around 60 were observed for both German-Upper Sorbian directions. However in the shared tasks run by AmericasNLP \cite{mager-etal-2021-findings}, nearly all BLEU scores were under 10, and in the ``full'' track of the 2021 Large-scale Multilingual shared task
\citep{wenzek2021wmt} the mean BLEU score of the best system was around 16, indicating low scores
for many of the 10,000 pairs. 

So, in terms of quantifiable results, it's a mixed picture with indications that translation can still be very poor in many cases, but if large-scale multilingual systems with ample synthetic data are possible, with a moderate amount of parallel data, then acceptable results can be obtained. However, even in the latter
circumstances (for Hausa-English) it's unclear how good the translations really are -- a detailed analysis
such as done by \citet{freitag_experts_2021} for German-English could be revealing. It is clear, however, that for languages pairs outside the 100--200 that have been considered so far, MT is likely to be very poor or even impossible due to the severe lack of data. This is especially true for non-English pairs, which are not much studied or evaluated in practice.



\smallskip

Looking forward, we now discuss a number of key areas for future work.

\paragraph{Collaboration with Language Communities}
Recent efforts by language communities~\cite{nekoto-etal-2020-participatory, martinus2020masakhane} to highlight their work and their lack of access to opportunities to develop expertise and contribute to language technology has resulted in bringing valuable talent and linguistic knowledge into our field. It is very clear that we need to
work together with speakers of low-resource and endangered languages to understand their challenges, their needs and to create knowledge and resources which can help them benefit from progress in our field, and also help their communities overcome language barriers. 

\paragraph{Massive Multilingual Models}
The striking success of multilingual pre-trained models such as mBART~\citep{liu2020multilingual} and mRASP~\cite{pan2021contrastive} still needs further investigation, and massively multilingual models clearly confer advantage to both high- and low-resource pairs~\citep{tran2021wmt,jian2021wmt}.  We should be able to answer questions such as whether the gains are more from the size of models, or from the number of languages the models are trained on, or is it from the sheer amount of data used. There are also questions about how to handle languages that are not included in the pretrained model. 

\paragraph{Incorporating external knowledge}
We will never have enough parallel data, and for many language pairs the situation is harder due to a lack of high-resourced related languages and a lack of monolingual data. We know that parallel data is not an efficient way to learn to translate. We have not fully explored questions such as what is a more efficient way of encoding translation knowledge --  bilingual lexicons, grammars or ontologies -- or indeed what type of knowledge is most helpful in creating MT systems and how to gather it. Further work looking at how we can best incorporate these resources is also needed: should they be incorporated directly into the model or should we use them to create synthetic parallel data? 

\paragraph{Robustness}
Modern MT systems, being large neural networks, tend to incur substantial quality degradation as the distribution of data encountered by the production system becomes more and more different than the distribution of the training data \citep{Lapuschkin2019, Hupkes2019, geirhos2020shortcut}. This commonly occurs in translation applications where the language domains, topics, and registers can be extremely varied and quickly change over time. Especially in low-resource settings, we are often limited to old training corpora from a limited sets of domains.
Therefore it is of great importance to find ways to make the systems robust to distribution shifts.
This is a big research direction in general machine learning, but it has a specific angle in MT due to the possibility of producing \emph{hallucinations} \citep{martindale-etal-2019-identifying, Raunak2021} that might mislead the user.
We need to find ways to make the systems detect out-of-distribution conditions and ideally avoid producing hallucinations or at least warn the user that the output might be misleading.

\smallskip

In conclusion, we hope that this survey will provide researchers with a broad understanding of low-resource machine translation, enabling them to be more effective at developing new tools and resources for this challenging, yet essential, research field.

\section*{Acknowledgements}

This work was partly funded by Rachel Bawden's chair in the PRAIRIE institute funded by the French national agency ANR as part of the ``Investissements d'avenir'' programme under the reference ANR-19-P3IA-0001.

\lettrine[image=true, lines=2, findent=1ex, nindent=0ex, loversize=.15]{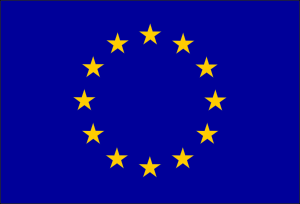}
This work has received funding from the European Union's Horizon 2020 Research and
Innovation Programme under Grant Agreement No 825299 (GoURMET). It was also supported by the UK Engineering and Physical Sciences Research Council fellowship grant EP/S001271/1 (MTStretch).

\starttwocolumn
\bibliographystyle{compling}
\bibliography{lowresource,WMT2021}

\end{document}